\documentclass[11pt]{article}

\usepackage[final]{acl}

\usepackage{times}
\usepackage{subcaption}
\usepackage{latexsym}
\usepackage{booktabs}
\usepackage{caption}
\usepackage[most]{tcolorbox}
\usepackage{fvextra}
\fvset{
  breaklines=true,       
  breakanywhere=false,   
  breaksymbol={},        
  breakindent=1.5em,     
  fontsize=\small,
}

\newtcolorbox{promptbox}[1]{
  colback=blue!4!white, colframe=blue!65!black,
  title={\textbf{#1}}, fonttitle=\small\bfseries,
  breakable, left=5pt, right=5pt, top=3pt, bottom=3pt,
}
\newtcolorbox{modelbox}[1]{
  colback=green!4!white, colframe=green!55!black,
  title={\textbf{#1}}, fonttitle=\small\bfseries,
  breakable, left=5pt, right=5pt, top=3pt, bottom=3pt,
}
\newtcolorbox{goldbox}[1]{
  colback=orange!5!white, colframe=orange!70!black,
  title={\textbf{#1}}, fonttitle=\small\bfseries,
  breakable, left=5pt, right=5pt, top=3pt, bottom=3pt,
}
\newtcolorbox{feedbackbox}[1]{
  colback=red!4!white, colframe=red!60!black,
  title={\textbf{#1}}, fonttitle=\small\bfseries,
  breakable, left=5pt, right=5pt, top=3pt, bottom=3pt,
}
\newtcolorbox{examplebox}[1]{
  colback=purple!4!white,
  colframe=purple!60!black,
  title={\textbf{#1}},
  fonttitle=\small\bfseries,
  breakable,
  left=5pt,
  right=5pt,
  top=3pt,
  bottom=3pt,
}

\usepackage[T1]{fontenc}

\usepackage[utf8]{inputenc}

\usepackage{microtype}

\usepackage{inconsolata}

\usepackage{graphicx}
\usepackage{enumitem} 

\newcommand{\paratitle}[1]{\vspace{1.0ex}\noindent\textbf{#1}}
\newcommand{\ie}{\textit{i.e.,}~}
\newcommand{\eg}{\textit{e.g.,}~}

\title{From Content Generation to Learning Support: Pedagogy-Guided Generative Video Tutors for STEM Learning}

\author{
\textbf{Xinchen Ma\textsuperscript{1}},
\textbf{Shuimu Wang\textsuperscript{1}},
\textbf{Gaole He\textsuperscript{2}\thanks{Corresponding authors.}},
\textbf{Yanbin Zhang\textsuperscript{1}},\\
\textbf{Chunyang Wang\textsuperscript{1}},
\textbf{Yunshi Lan\textsuperscript{1}\footnotemark[1]},
\textbf{Weining Qian\textsuperscript{1}} \\
\textsuperscript{1}Department of Data Science and Engineering, East China Normal University \\
\textsuperscript{2}School of Computing, National University of Singapore \\
\texttt{xinchen.ma@stu.ecnu.edu.cn},
\texttt{hegaole@nus.edu.sg},
\texttt{yslan@dase.ecnu.edu.cn}
}

\begin{document}
\maketitle
\begin{abstract}
Generative AI enables scalable production of educational videos, but current systems largely focus on producing visually coherent content rather than supporting learning. As a result, generated videos often lack explicit pedagogical structure, reliable quality control, and mechanisms for assessing learner understanding or addressing misconceptions. 
In this work, we introduce \textbf{PIVOT} (\textbf{P}edagogy-guided \textbf{I}nstructional \textbf{V}ideO \textbf{T}utoring), 
a generative video tutoring framework for STEM learning via learning-centered instructional support.\footnote{Code is available at
\href{https://github.com/Chloe-mxxxxc/Pedagogy-Guided-Generative-Video-Tutors-for-STEM-Learning}
{our GitHub repository}.}
Inspired by conventional teaching workflows, our framework integrates pedagogy into the full generation pipeline: it first uses instructional principles to guide storyboard generation, then produces verified multimodal videos through code-centric generation and a pedagogical verification harness, and finally connects videos with assessment and misconception-aware remediation. 
Experiments and expert evaluations across four STEM domains show that our framework produces educational videos with pedagogically aligned content, clear and engaging presentation, coherent instructional flow, and perceived effectiveness for learning.
These findings suggest a human-centered perspective on educational content generation: generative systems should be evaluated and designed not only by what they produce, but also by how they support teaching practices, learner understanding, and corrective feedback.


\end{abstract}

\section{Introduction}
Preparing instructional materials and recording lectures is a time-consuming task for educators, and providing adaptive feedback to support multiple students can be an additional burden~\cite{shi2025educationq}.
The rise of generative AI has opened new opportunities for creating high-quality educational videos that can foster student engagement and facilitate understanding ~\cite{ye2025position,pal2024autotutor,graesser2004autotutor}.
Recent work has explored generating visually coherent educational videos from learning goals~\cite{chen2025code2video,ku2025theoremexplainagent}, demonstrating the feasibility of translating instructional intents into multimodal learning materials at scale.
While these systems ensure content fidelity and visual fluency, they primarily focus on content generation rather than learning support. 
Consequently, several important aspects of educational video design remain largely unaddressed~\cite{hu2025generative,zerkouk2025comprehensivereviewaibasedintelligent}.
First, videos often lack explicit \textbf{pedagogical guidance}, such as clear learning objectives, prerequisite activation, and structured explanations, which can make it difficult for learners to follow and internalize new knowledge. Second, ensuring that visualizations, narration, and examples are \textbf{factually correct and cognitively clear} remains challenging, as errors or ambiguous layouts can impede learning or lead to misconceptions. 
Third, existing approaches typically do not incorporate \textbf{formal assessment of learning outcomes} or mechanisms for \textbf{corrective feedback}, limiting insight into whether students have grasped the intended learning goals. 
Together, these limitations highlight gaps in current generative video research regarding pedagogical support, content quality, and learning-centered evaluation.

To address these challenges, we introduce \textbf{PIVOT}, a three-stage pedagogy-guided generative video tutoring framework for STEM education that supports active learning. 
The first two stages build upon prior code-centric video generation frameworks~\cite{chen2025code2video,ku2025theoremexplainagent,chen2025visualedu} and introduce two key improvements. First, pedagogical guidance is incorporated during the instructional storyboard generation. The system structures learning objectives, activates prerequisite knowledge, and provides step-by-step explanations along with visual representation plans and worked examples. This approach ensures that the instructional content is aligned with established principles of instructional design, scaffolding, and multimedia learning. ~\cite{cohn2026theory,baek2026pedaco,mayer2005cognitive}
Second, video quality is enhanced through a pedagogical verification harness, ensuring that rendered animations are factually correct, visually clear, and pedagogically aligned, thereby improving the robustness and interpretability of the generated instructional videos.

In the final stage, learner assessment is incorporated to evaluate learning outcomes and address misconceptions. 
For each quiz question derived from the instructional storyboard, the system provides feedback on all answer options, while incorrect responses trigger a targeted remediation video that contrasts the incorrect choice with the correct reasoning.
These contrastive explanations clarify underlying misconceptions and reinforce new knowledge, providing tailored guidance that addresses each learner’s specific errors. 
By dynamically generating instructional content in response to learner performance, this stage connects AI-generated videos with post-video formative assessment. 
In this way, the framework shifts educational video generation from passive content delivery toward active learning support and learner-centered instructional interaction.



Extensive evaluation demonstrates that the proposed PIVOT framework can produce educational videos that effectively support STEM learning, ensuring both high-quality instructional content and meaningful learner assessment. Expert evaluation conducted with experienced high school and university instructors confirms that the videos are pedagogically sound and facilitate conceptual understanding. The contributions of this work can be summarized as follows:

\begin{itemize}[leftmargin=12pt, topsep=0pt, partopsep=0pt, parsep=0pt, itemsep=1pt]
    \item We introduce \textbf{PIVOT}, a pedagogy-guided instructional video tutoring framework that integrates storyboard-based planning, verified multimodal video generation, and assessment-driven remediation to support STEM learning.
    \item We validate the framework through experiments and expert assessments, demonstrating that it reliably generates videos that support student comprehension and address misconceptions.
    \item The framework represents a step toward AI-supported STEM learning guided by pedagogy, transforming AI-generated videos from passive content into interactive, pedagogically informed learning experiences.
    
    
\end{itemize}
\begin{figure*}[htbp]
\centering
\includegraphics[width=0.9\textwidth]{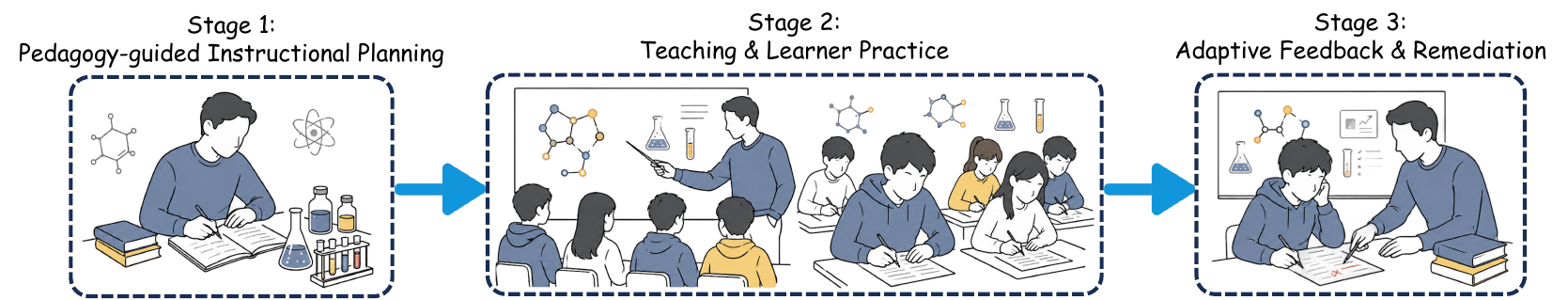}
\caption{Real-world instructional workflow.}
\label{fig:teaser}
\end{figure*}

\section{Related Work}


\subsection{Text-to-Video Generation}
Recent advances in text-to-video generation models, such as Sora ~\cite{liu2024sora}, Kling ~\cite{team2025kling}, and other large-scale generative video systems ~\cite{veo3_2025,zheng2025open,bar2024lumiere}, have substantially expanded the research scope of AI-assisted education by enabling automatic synthesis of dynamic visual content from textual descriptions. These systems demonstrate strong capabilities in cinematic video synthesis and multimodal storytelling. However, applying general-purpose video generation models to STEM education remains challenging due to issues in temporal consistency, layout stability, and accurate rendering of symbolic content such as formulas, diagrams, and algorithms~\cite{chen2024videocrafter2}. Such limitations reduce the pedagogical reliability required for instructional videos.

\subsection{Code-centric Educational Video Generation}

Early LLM-based (Large Language Model) educational applications primarily focused on language-centric tasks, including question answering, exercise generation, and conversational tutoring ~\cite{mageira2022educational,grassini2023shaping,lee2023lecture,zhang2025simulating,bi2026eduillustrate,baek2026pedaco,dickey2024gaide}. With the rapid development of multimodal generation capabilities, recent research has gradually shifted toward richer forms of educational content generation, including visual explanations, instructional animations, and educational videos. 


\subsection{Pedagogical Guidance and Adaptive Educational Generation}
Recent research in Intelligent Tutoring Systems (ITS) and AI-assisted education has explored various forms of pedagogical guidance ~\cite{pal2024autotutor,borchers2025can,schmucker2024ruffle,graesser2004autotutor,alshaikh2024implementation,calo2024towards,yang2025llm}, including adaptive learning, learner modeling, formative assessment, corrective feedback, and scaffolded instructional support~\cite{zerkouk2025comprehensivereviewaibasedintelligent,cohn2026theory,ye2025position}.
Prior work has shown that pedagogically informed tutoring systems can improve learner engagement and conceptual understanding through adaptive instruction based on learner performance.

Despite recent advances, existing educational video generation systems mainly focus on multimedia synthesis with limited support for pedagogical planning, assessment, and misconception-aware feedback. In contrast, our framework integrates pedagogical guidance, formative assessment, and remediation into a unified pipeline, transforming educational video generation into learner-centered instructional interaction.


\section{PIVOT: Pedagogy-guided Instructional VideO Tutoring}

\subsection{System Architecture}
Designing effective educational videos demands careful alignment with pedagogical principles, including lesson planning, instructional delivery, learner practice, and adaptive feedback. 
In conventional STEM education, instructors typically follow a structured workflow (as shown in Figure~\ref{fig:teaser}): they first prepare lesson plans that organize instructional goals, prerequisite knowledge, worked examples, and assessments. They then present lessons to students through demonstrations or lectures. Students engage in exercises to consolidate understanding, and instructors provide targeted feedback to address misconceptions and reinforce learning.

\begin{figure*}[t]
\centering
\includegraphics[width=0.93\textwidth]{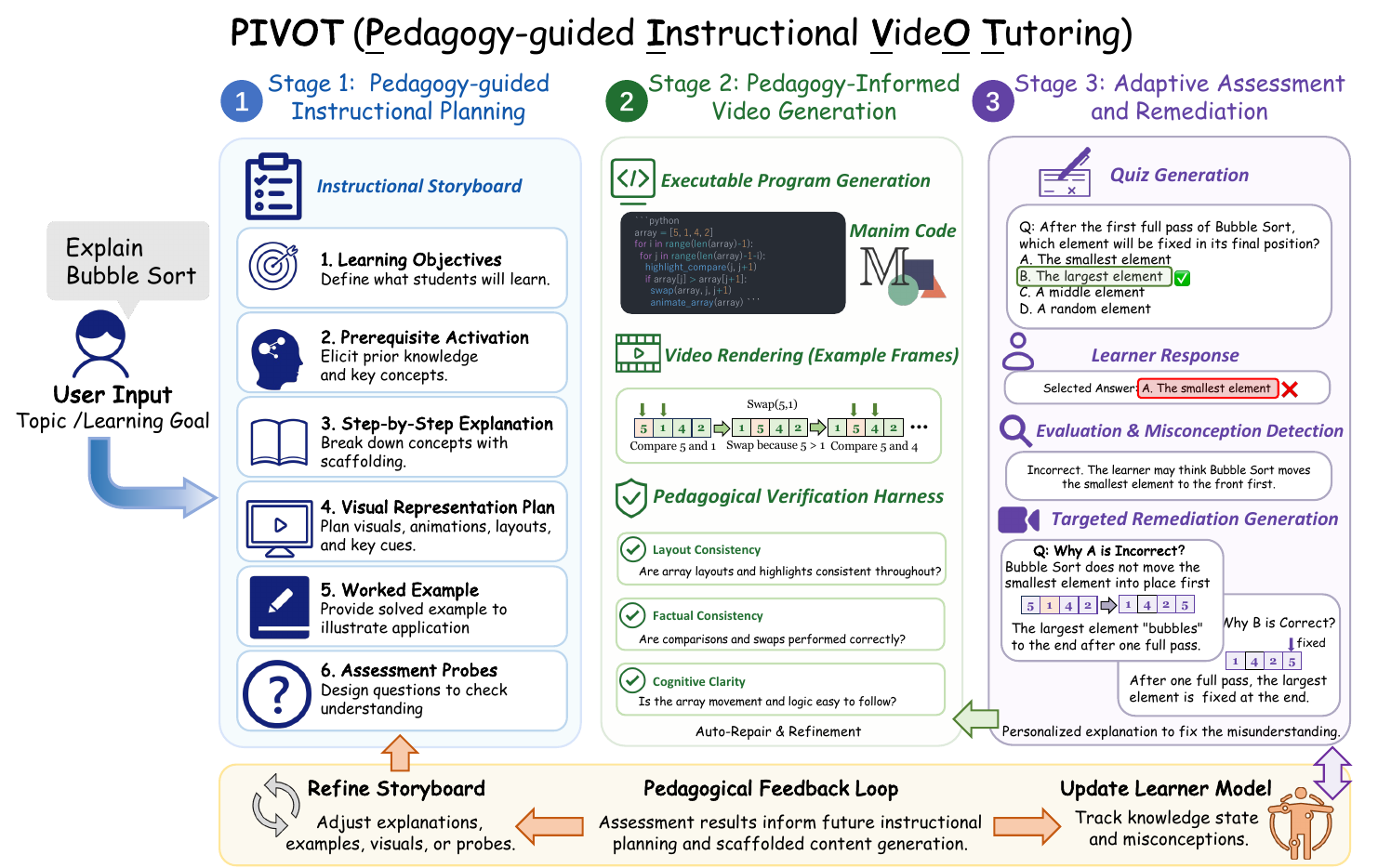}
\vspace{-0.3em}
\caption{Overview of the PIVOT Framework.}
\label{fig:method}
\end{figure*}

Inspired by this instructional process, our framework is structured into three stages (shown in Figure~\ref{fig:method}).
The first stage, \textbf{Pedagogy-guided Instructional Planning}, emulates lesson preparation by generating structured storyboards that organize learning objectives, prerequisite activation, stepwise explanations, visual representation plans, worked examples and formative assessment probes. 
This stage ensures that educational content is pedagogically coherent and scaffolded, providing a solid foundation for downstream video generation.
The second stage, \textbf{Video Generation with Pedagogical Verification Harness}, transforms the storyboard into shot-level animations. 
Building upon prior code-centric video generation frameworks~\cite{chen2025code2video,ku2025theoremexplainagent}, this stage integrates a multimodal verification harness to improve visual clarity, factual correctness, and cognitive clarity, enhancing the quality and interpretability of generated videos.
The third stage, \textbf{Adaptive Assessment and Remediation}, parallels post-class student practice and feedback, generating quizzes and targeted explanatory videos to identify misunderstandings and reinforcing conceptual understanding.
By integrating pedagogy across the pipeline, our framework turns educational video generation from passive content production into learner-centered support.


\subsection{Pedagogy-guided Instructional Planning}

To generate high-quality educational videos, 
we first integrate pedagogical principles into the creation of instructional materials. 
Simply generating video frames without structured content risks producing visually coherent but educationally ineffective material. 
By constructing a structured instructional storyboard, we ensure that content is not only organized but also \textbf{pedagogically meaningful}, providing learners with a clear path through complex concepts. 
This intermediate representation allows the system to embed instructional scaffolding, manage cognitive load, and prepare content that is interpretable by downstream video generation modules, ultimately guiding the learner toward progressively deeper understanding.


Given a learner-specified topic, the storyboard includes components grounded in educational theory, each serving a clear pedagogical purpose:
\begin{itemize}[leftmargin=12pt, topsep=0pt, partopsep=0pt, parsep=0pt, itemsep=1pt]
    \item \textbf{Learning Objectives}: Define the target concepts and expected outcomes, clarifying the purpose of each lesson and focusing both the content generation and the learner’s attention~\cite{bloom1956taxonomy}.
    
    \item \textbf{Prerequisite Activation}: Recall relevant prior knowledge to reduce cognitive load and support integration of new concepts, preparing learners to grasp advanced material~\cite{sweller1988cognitive}.
    
    \item \textbf{Concept Explanations}: Present ideas progressively through stepwise reasoning, scaffolding complex concepts and enabling incremental knowledge acquisition~\cite{wood1976role,cohn2026theory}.
    
    \item \textbf{Visual Representation Plans}: Specify diagrams, formulas, highlights, and animations according to multimedia learning principles, enhancing comprehension, retention, and cognitive clarity~\cite{mayer2005cognitive}.
    
    \item \textbf{Worked Examples}: Demonstrate procedural steps and intermediate reasoning, helping learners internalize problem-solving strategies and reinforcing conceptual understanding~\cite{renkl2014toward}.
    
    \item \textbf{Diagnostic Assessment Probes}: Include formative questions to detect misconceptions early, guiding adaptive feedback and reinforcing learning~\cite{black1998assessment}
\end{itemize}

This careful preparation helps bridge educational theory and AI-generated content, laying a solid foundation for producing videos that are coherent, scaffolded, and pedagogically effective. 
The generated storyboard further serves as the semantic backbone for executable Manim\footnote{\url{https://github.com/ManimCommunity/manim}} code generation, narration synthesis, assessment construction, and multimodal verification~\footnote{Examples of the storyboard are provided in Appendix~\ref{appendix:instructional_planning_examples}.}.

\subsection{Video Generation with Pedagogical Verification Harness}

Building on the storyboard from Stage 1, the system transforms structured instructional content into shot-level executable videos. Each shot contains pedagogically grounded components, including local learning goals, narration, visual descriptions, visual elements (\eg formulas, diagrams, highlights), and duration constraints. 
Dividing the content into shot-level units allows precise control over each instructional segment, facilitates iterative verification and repair, and ensures that pedagogical objectives are faithfully represented. 
By using this intermediate representation, the system ensures that videos preserve instructional structure and scaffolded learning.

\begin{table}[htbp]
\centering
\small
\setlength{\tabcolsep}{2pt}
\begin{tabular}{p{0.33\linewidth} p{0.60\linewidth}}
\toprule
\textbf{Component} & \textbf{Function} \\
\midrule

Layout Consistency
& Detects overlap, spacing inconsistency, readability issues, and excessive information density.
\\
\midrule
Factual Consistency
& Verifies formulas, symbolic transitions, algorithmic procedures, and instructional explanations.
\\
\midrule
Cognitive Clarity 
& Verifies logical consistency, instructional progression, and coherence of explanatory content.
\\

\bottomrule
\end{tabular}
\caption{Pedagogical verification components.}
\label{tab:verification_harness_full}
\end{table}

\paratitle{Pedagogical Verification Harness}. 
To ensure that generated videos faithfully implement the instructional storyboard, we employ a multimodal verification harness. The harness is designed to detect potential issues that could compromise educational fidelity or visual clarity, covering layout, cognitive clarity and factual consistency. As summarized in Table~\ref{tab:verification_harness_full}, each component guides iterative repair, ensuring that pedagogical intent is preserved while improving interpretability and robustness.

\paratitle{Verification-guided Rendering and Repair.} 
To further maintain instructional and visual quality, the system applies constrained layout prompting (see Appendix ~\ref{appendix:Constrained Layout Prompt}) and automated code repair.
These mechanisms reduce common generation failures such as overlapping elements, unreadable text, and inconsistent scene transitions, preserving the instructional structure of each shot. 
When verification failures are detected, the generated Manim code is iteratively repaired using low-temperature prompting conditioned on execution traces and rendering feedback. GPT-5.2 serves as the primary generation and repair model, while persistent failures are transferred to Claude Sonnet 4.6 for additional debugging and executable repair.

After verification, a Text-to-Speech (TTS) module generates synchronized narration aligned with the rendered animations. 
For remediation videos, generation is further conditioned on learner responses (\ie quiz questions, candidate options, incorrect answers, and corresponding correct reasoning), enabling adaptive, misconception-aware explanations.
By grounding video generation in the pedagogically structured storyboard and enforcing multi-level verification, this stage ensures that the generated videos remain aligned with the intended instructional objectives.

\subsection{Adaptive Assessment and Remediation}

To provide learner-centered feedback, address misconceptions, and reinforce conceptual understanding, the framework incorporates an adaptive assessment and remediation module. This module generates quizzes and targeted follow-up explanations based on the instructional storyboard from Stage 1, forming a closed feedback loop that aligns instructional content, learner responses, and pedagogical objectives while actively addressing misunderstandings.

\paratitle{Pedagogically Grounded Quiz Generation.} Quiz generation leverages the structured storyboard, summarizing shot-level content including learning objectives, explanations, key concepts, worked examples, and visual demonstrations. 
This design ensures that generated questions remain semantically aligned with the instructional video and intended learning goals.
To encourage conceptual understanding rather than memorization, our framework does not reuse examples, equations, or numerical values from the generated educational video. 
By grounding assessments in the pedagogical structure of the lesson, the framework ensures alignment with instructional goals and supports learners in building robust conceptual knowledge.

\paratitle{Misconception-aware Remediation Generation.} Learner responses provide adaptive pedagogical signals to guide remediation. When a learner answers incorrectly, the system generates a targeted explanatory video contrasting the incorrect reasoning with the correct solution. The remediation generation is conditioned on the quiz question, candidate options, learner response, and the correct answer, ensuring that feedback addresses specific misconceptions. By integrating learner signals with the instructional storyboard, the framework produces remediation that preserves semantic consistency, instructional fidelity, and pedagogical coherence, supporting learners in correcting misunderstandings and reinforcing knowledge.

Through adaptive pedagogy-guided generation, the system connects instructional content, video synthesis, and learner interaction to support effective learning.




\section{Experiments}

For evaluation, we select two representative baseline systems: \textbf{TheoremExplainAgent}~\cite{ku2025theoremexplainagent}, and \textbf{Code2Video}~\cite{chen2025code2video}. 
These systems represent state-of-the-art approaches in automatic educational video generation, primarily focusing on content synthesis without explicit pedagogical guidance or learner-adaptive feedback. To ensure fair comparison, all methods are evaluated on the same set of 40 STEM instructional topics spanning mathematics, physics, chemistry, and computer science, with 10 knowledge points selected for each subject (see Appendix ~\ref{Instructional Topics Used for Video Generation and Evaluation}).

\subsection{Automatic Evaluation}

To evaluate generated instructional videos, we conduct automatic assessment using large Vision-Language Models (VLMs). Video frames sampled from the central timeline are independently evaluated by Gemini-2.5-Flash~\cite{imran2024google} under deterministic decoding ($T=0$). Each frame is scored on two dimensions: \textbf{\textit{Layout}}, measuring spatial clarity and visual organization, and \textbf{\textit{Richness}}, evaluating visual diversity and instructional content coverage. These dimensions primarily evaluate visual presentation quality, while higher-level pedagogical effectiveness is assessed separately through human evaluation. In Table~\ref{tab:vlm_video_eval}, final video scores are obtained by averaging frame-level results across all sampled frames.

\begin{table}[htbp]
\centering
\small
\setlength{\tabcolsep}{6pt}
\scalebox{.98}{
\begin{tabular}{lccc}
\toprule
\textbf{Method} & \textbf{Layout} & \textbf{Richness} & \textbf{Overall} \\
\midrule
TheoremExplainAgent & 9.00 & 4.23 & 6.62 \\
Code2video &  8.18& 4.22  & 6.20 \\
PIVOT & \textbf{9.10} & \textbf{6.06} & \textbf{7.58} \\
\bottomrule
\end{tabular}}
\caption{Video quality evaluation results (1--10).}
\label{tab:vlm_video_eval}
\end{table}

In Table~\ref{tab:transcript_eval}, we further evaluate instructional narration quality using GPT-4o~\cite{openai2024gpt4technicalreport}. Narration fields are extracted from generated scripts and independently scored along two pedagogically grounded dimensions: \textbf{\textit{Accuracy and Depth}}, assessing conceptual correctness and explanatory quality, and \textbf{\textit{Logical Flow}}, evaluating instructional coherence and progression. The final transcript score is computed as their arithmetic mean~\footnote{Detailed evaluation prompts are provided in Appendix~\ref{Automatic Evaluation}.}.
Overall, our method achieves the best performance on both video and narration quality evaluation, demonstrating improved visual organization, instructional richness, conceptual accuracy, and logical coherence.

\begin{table}[htbp]
\centering
\small
\setlength{\tabcolsep}{6pt}
\scalebox{.98}{
\begin{tabular}{lccc}
\toprule
\textbf{Method} & \textbf{Accuracy} & \textbf{Logical} & \textbf{Overall} \\
\midrule
TheoremExplainAgent & 0.85 & 0.82 & 0.84 \\
Code2video &  0.78& 0.72 & 0.75  \\
PIVOT & \textbf{0.87} & \textbf{0.91} & \textbf{0.89} \\
\bottomrule
\end{tabular}}
\caption{Narration quality evaluation results (0--1).}
\label{tab:transcript_eval}
\end{table}

\begin{figure*}[t]
\centering

\begin{subfigure}[t]{0.48\textwidth}
    \centering
    \includegraphics[width=\linewidth]{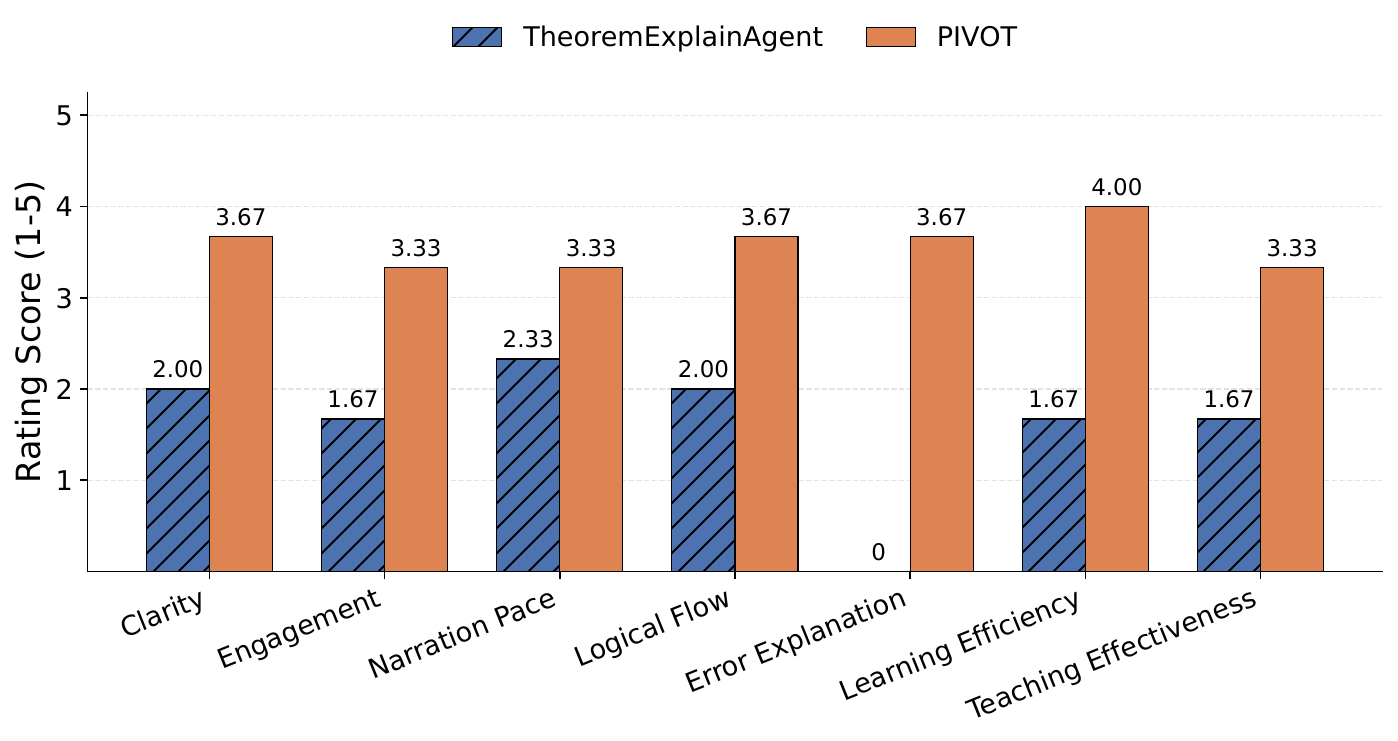}
    \caption{Physics}
\end{subfigure}
\hfill
\begin{subfigure}[t]{0.48\textwidth}
    \centering
    \includegraphics[width=\linewidth]{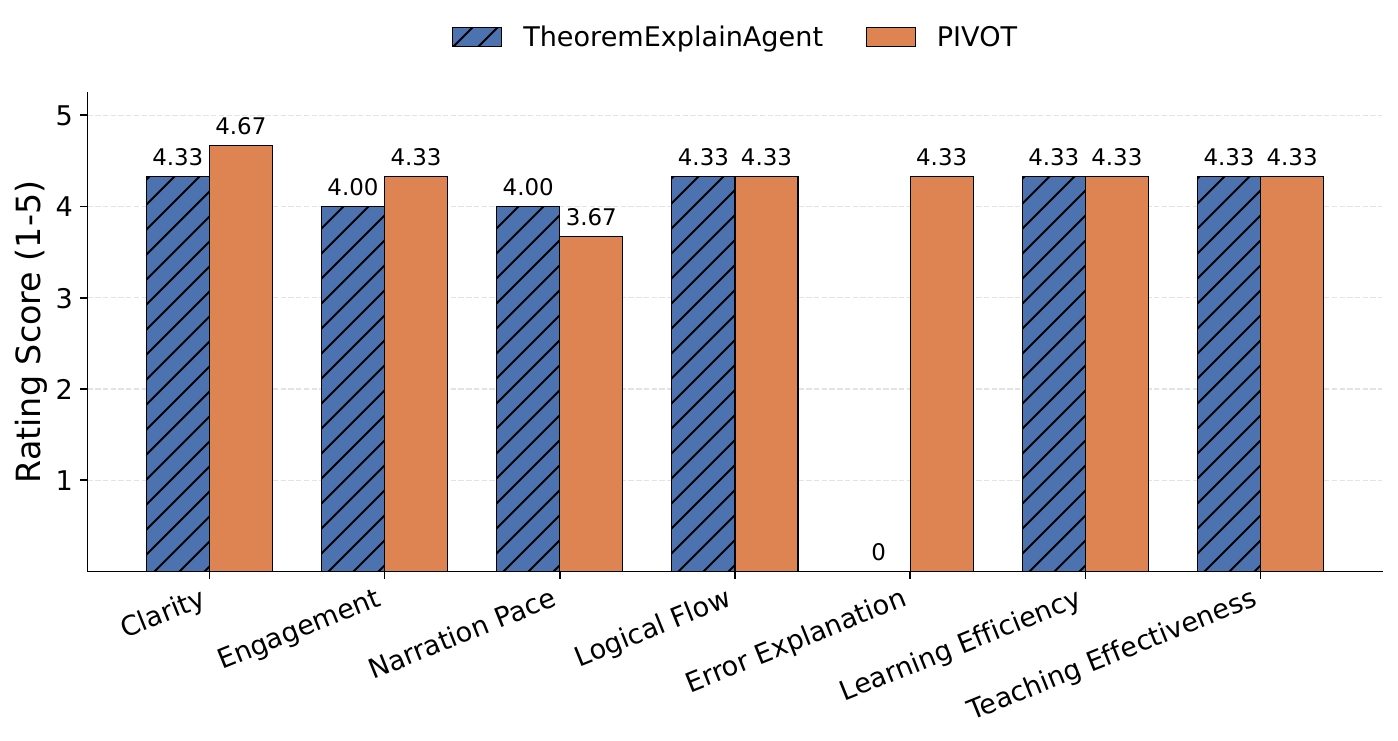}
    \caption{Chemistry}
\end{subfigure}

\vspace{-0.2em}

\begin{subfigure}[t]{0.48\textwidth}
    \centering
    \includegraphics[width=\linewidth]{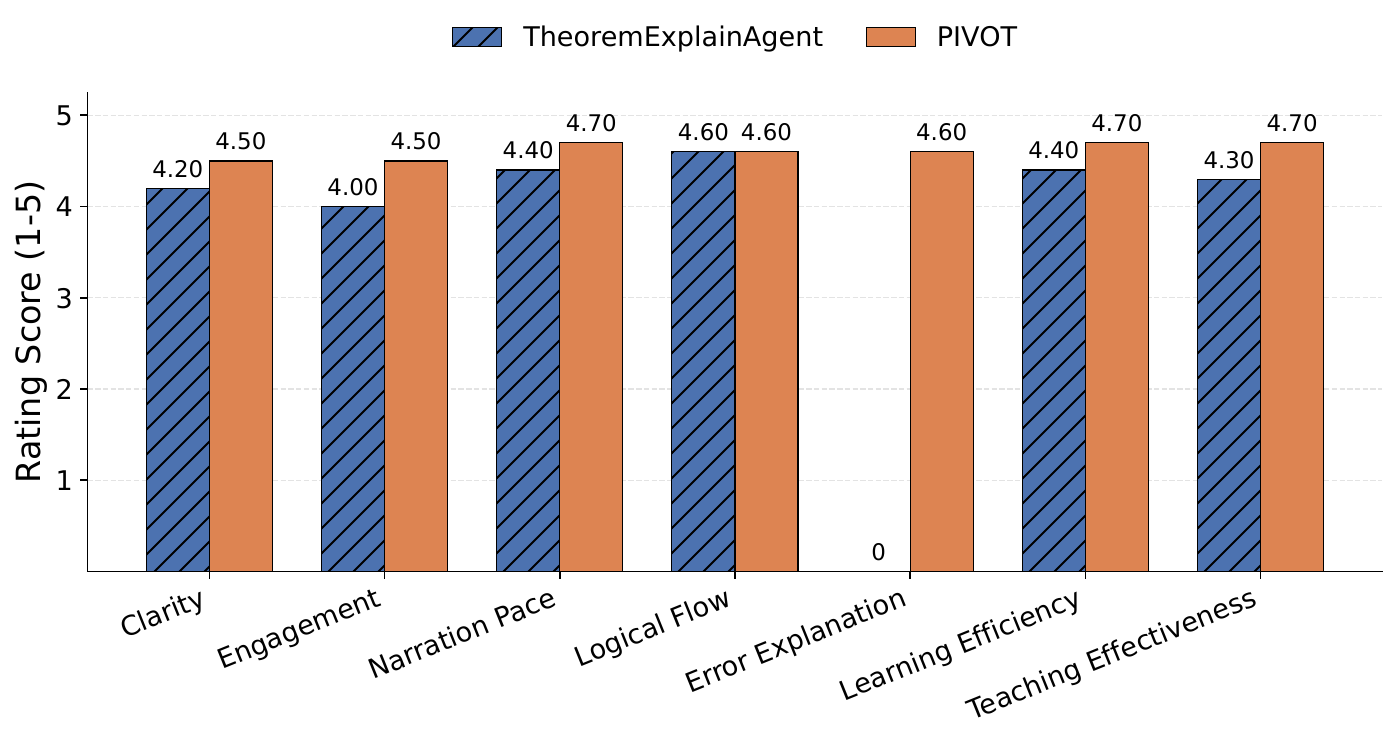}
    \caption{Mathematics}
\end{subfigure}
\hfill
\begin{subfigure}[t]{0.48\textwidth}
    \centering
    \includegraphics[width=\linewidth]{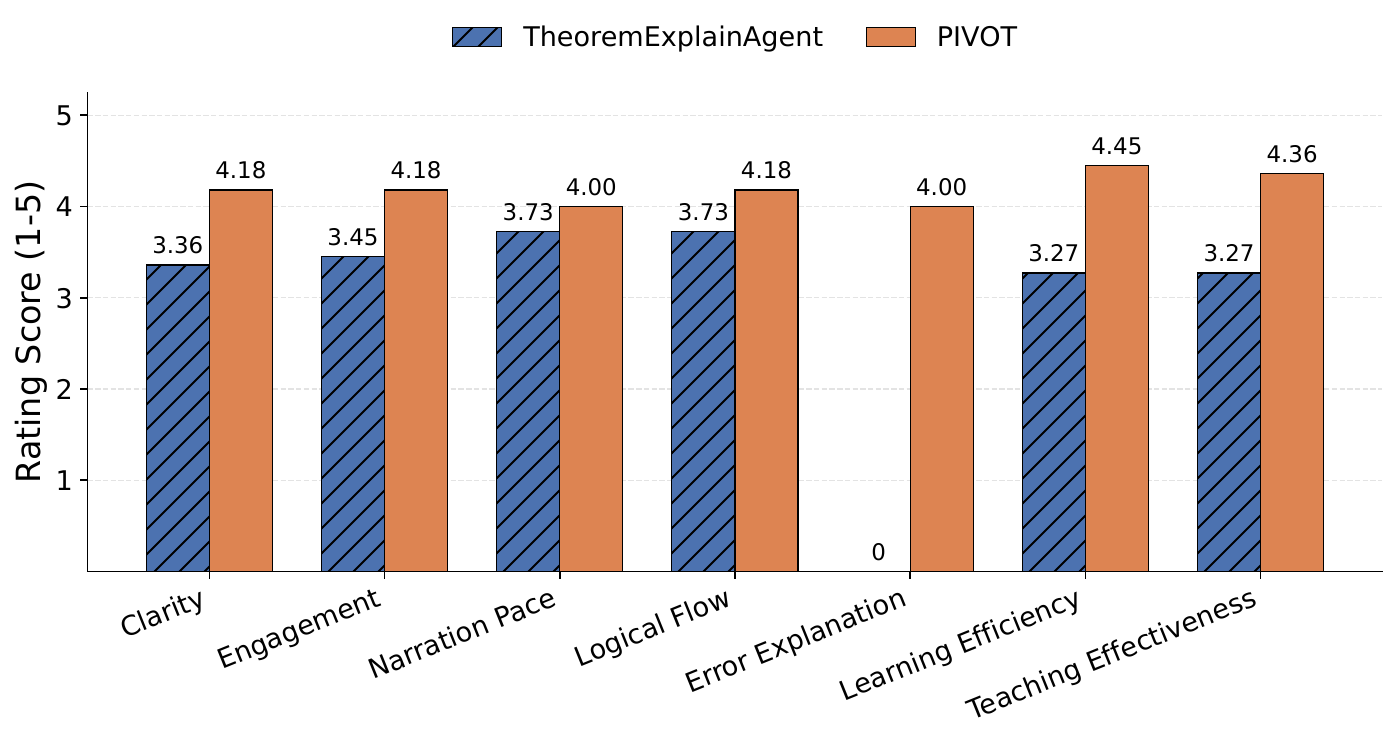}
    \caption{Computer Science}
\end{subfigure}

\vspace{-0.2em}
\caption{
Expert evaluation results across four STEM domains. }

\label{fig:user_study_bar}
\end{figure*}


\subsection{Ablation Analysis}

We further conduct an ablation study to analyze the contribution of three core verification components in the proposed framework: \textbf{Layout Consistency}, \textbf{Factual Consistency}, and \textbf{Cognitive Clarity }.
As summarized in Table~\ref{tab:ablation_metrics}, these modules mainly affect visual organization, conceptual correctness, and instructional coherence, respectively. We evaluate their impact on both visual quality and narration quality.


\begin{table}[htbp]
\centering
\small
\setlength{\tabcolsep}{4pt}
\renewcommand{\arraystretch}{1}

\begin{tabular}{lccc}
\toprule
\textbf{Ablation} 
& \textbf{Layout} 
& \textbf{Accuracy} 
& \textbf{Logical}
\\
\midrule

w/o Layout Verification
& 7.87 & -- & --
\\

w/o Factual Verification
& -- & 0.85 & --
\\

w/o Cognitive Verification
& -- & -- & 0.89
\\

Full Framework (PIVOT)
& 9.10 & 0.87 & 0.91
\\

\bottomrule
\end{tabular}

\caption{Ablation study of core verification components in the proposed framework. }
\label{tab:ablation_metrics}
\end{table}

First, removing the \textbf{Layout Consistency} module significantly degrades the Layout score from 9.10 to 7.87. This result demonstrates that 
layout verification plays an important role in maintaining spatial clarity and visual cleanliness. Without explicit layout control, generated scenes are more likely to contain overlapping elements and inconsistent organization, reducing instructional readability. This module mainly affects visual quality rather than narration generation.

Second, removing \textbf{Factual Consistency} decreases the Accuracy \& Depth score from 0.87 to 0.85, indicating that factual verification improves conceptual correctness and explanatory reliability. Without this module, generated narrations are more likely to contain imprecise explanations or incomplete reasoning. For example, in chemical equation balancing, the model may incorrectly describe the law of conservation of mass as conservation of atom counts, potentially introducing learner misconceptions.

Third, removing the \textbf{Cognitive Clarity} module decreases the Logical Flow score from 0.91 to 0.89. The resulting narrations are more likely to exhibit abrupt topic transitions, weaker instructional progression, and less coherent explanation structures. This finding indicates that explicit logical verification improves instructional coherence and helps the generated videos better follow pedagogically structured teaching sequences rather than mechanically listing isolated reasoning steps.

\begin{figure*}[t]
\centering

\begin{minipage}[t]{0.32\textwidth}
    \centering
    \includegraphics[
        width=\linewidth,
        height=0.115\textheight
    ]{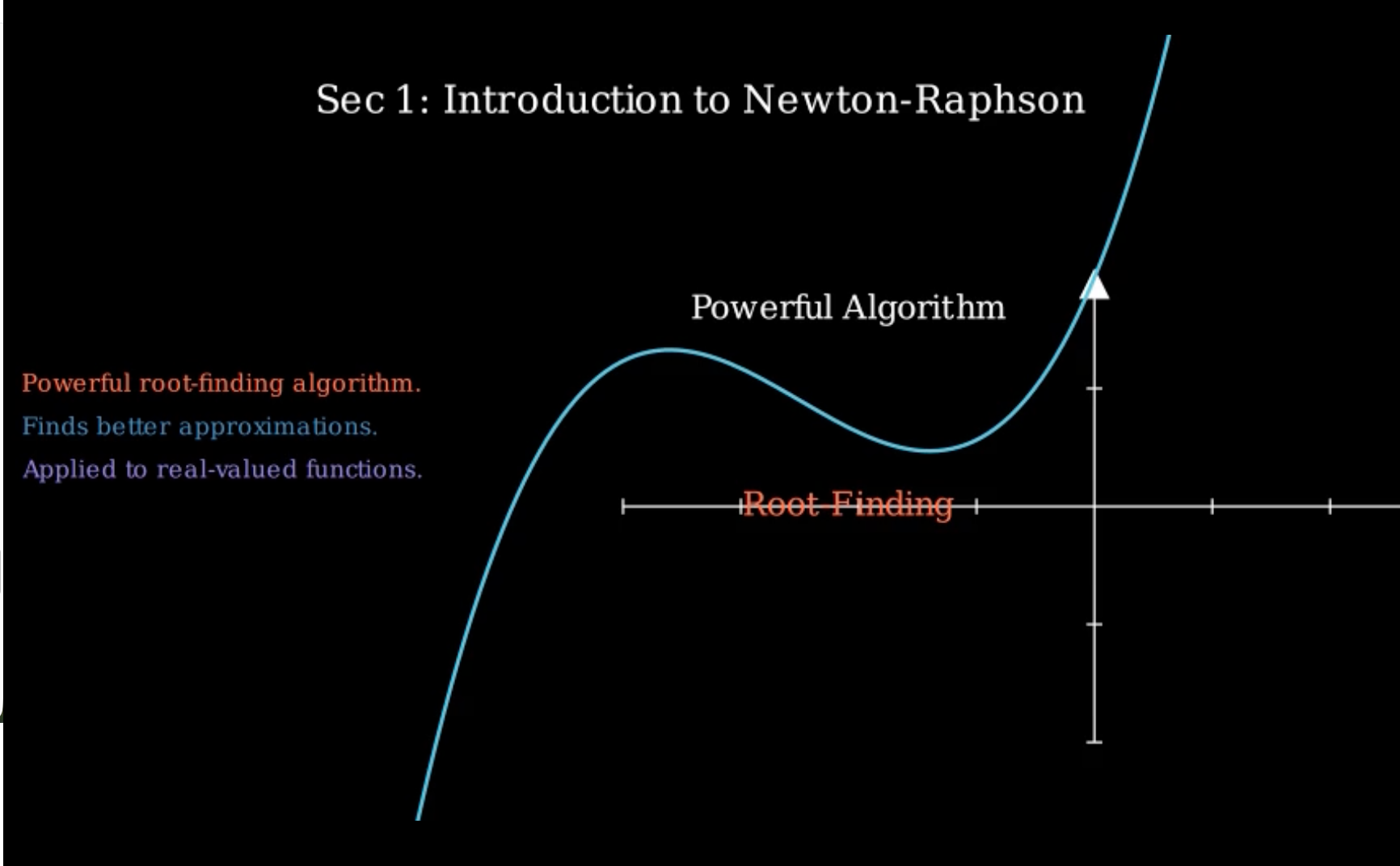}
\end{minipage}
\hfill
\begin{minipage}[t]{0.32\textwidth}
    \centering
    \includegraphics[
        width=\linewidth,
        height=0.115\textheight
    ]{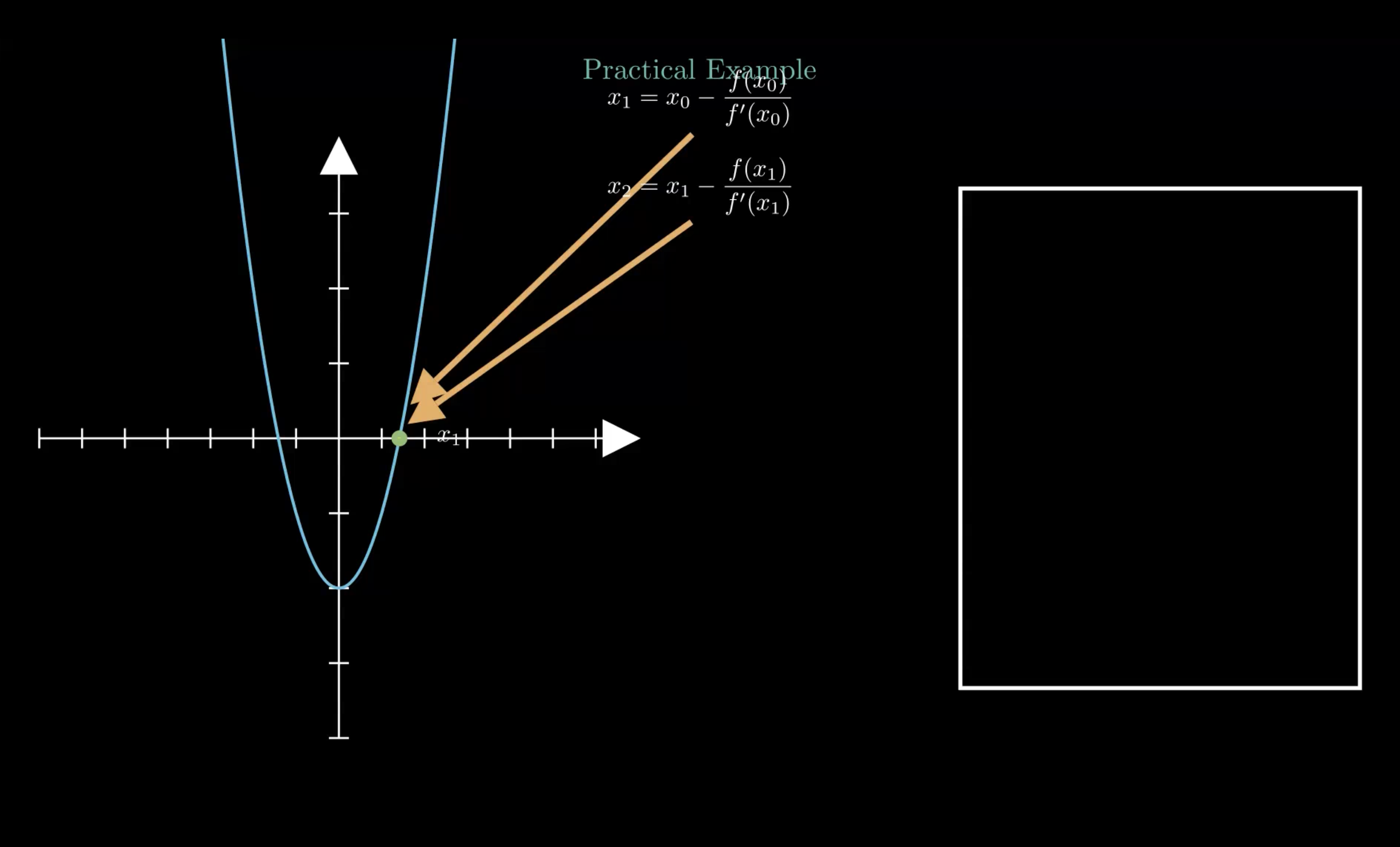}
\end{minipage}
\hfill
\begin{minipage}[t]{0.32\textwidth}
    \centering
    \includegraphics[
        width=\linewidth,
        height=0.115\textheight
    ]{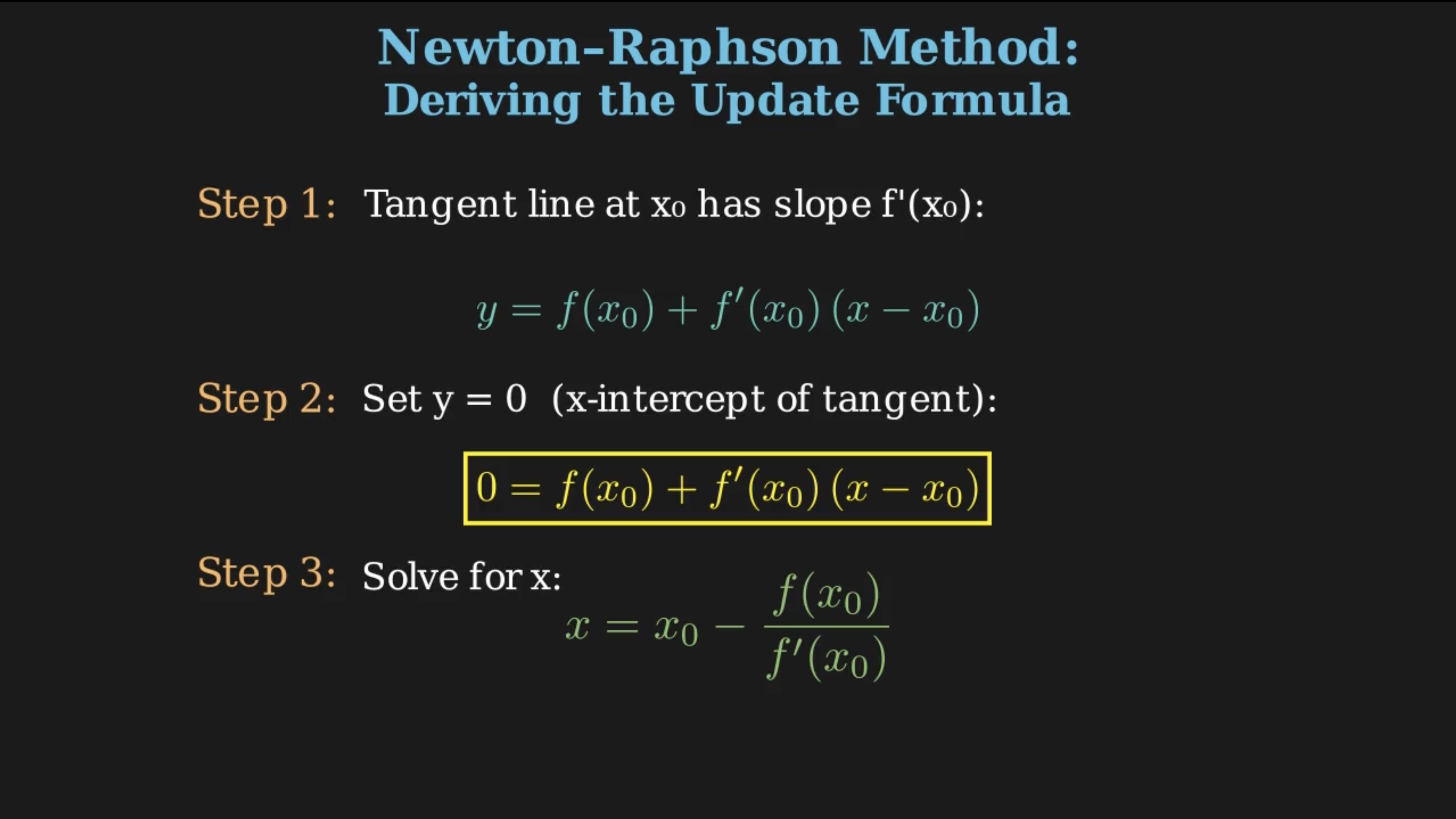}
\end{minipage}

\vspace{0.05em}

\begin{minipage}[t]{0.32\textwidth}
    \centering
    \includegraphics[
        width=\linewidth,
        height=0.115\textheight
    ]{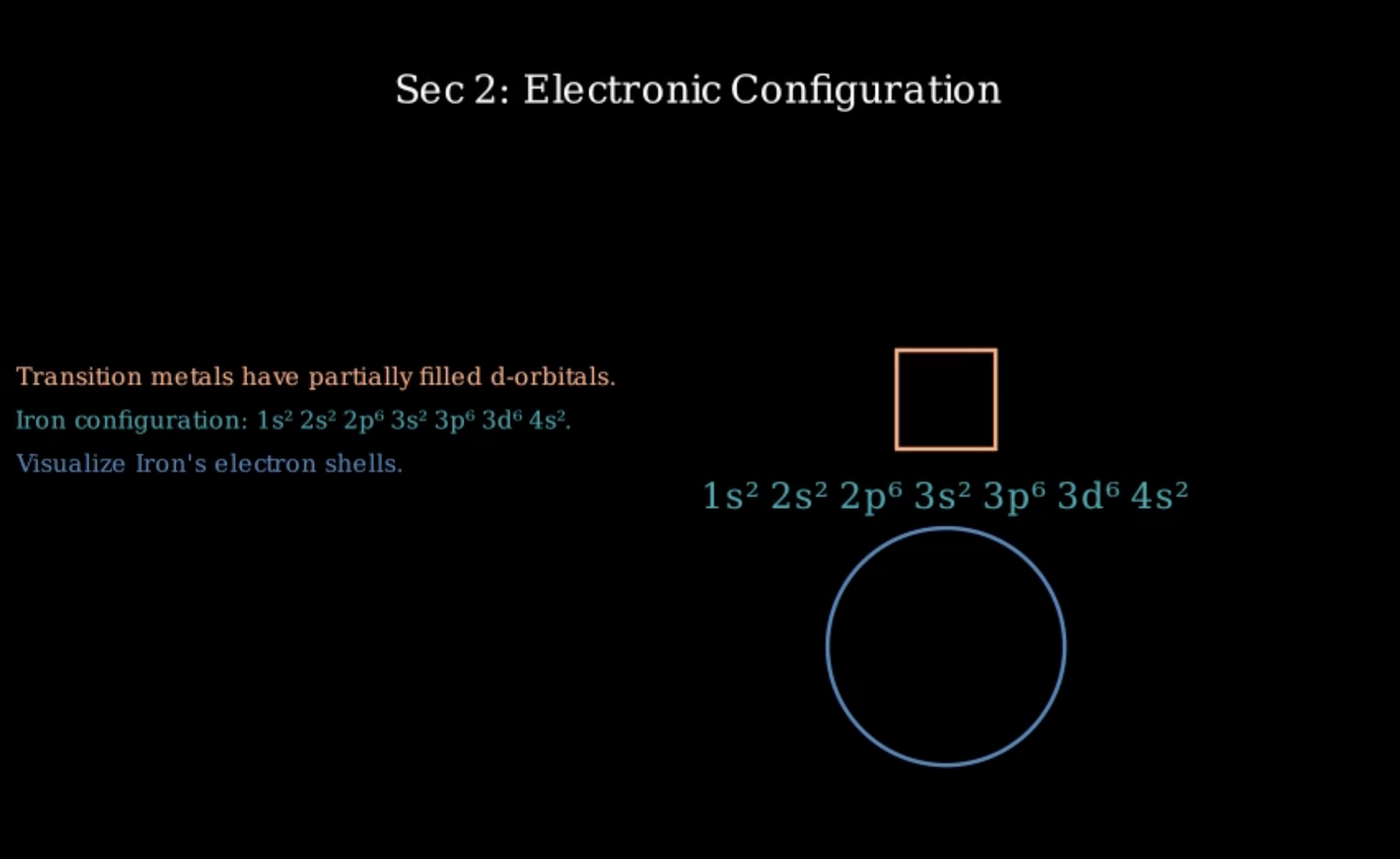}
\end{minipage}
\hfill
\begin{minipage}[t]{0.32\textwidth}
    \centering
    \includegraphics[
        width=\linewidth,
        height=0.115\textheight
    ]{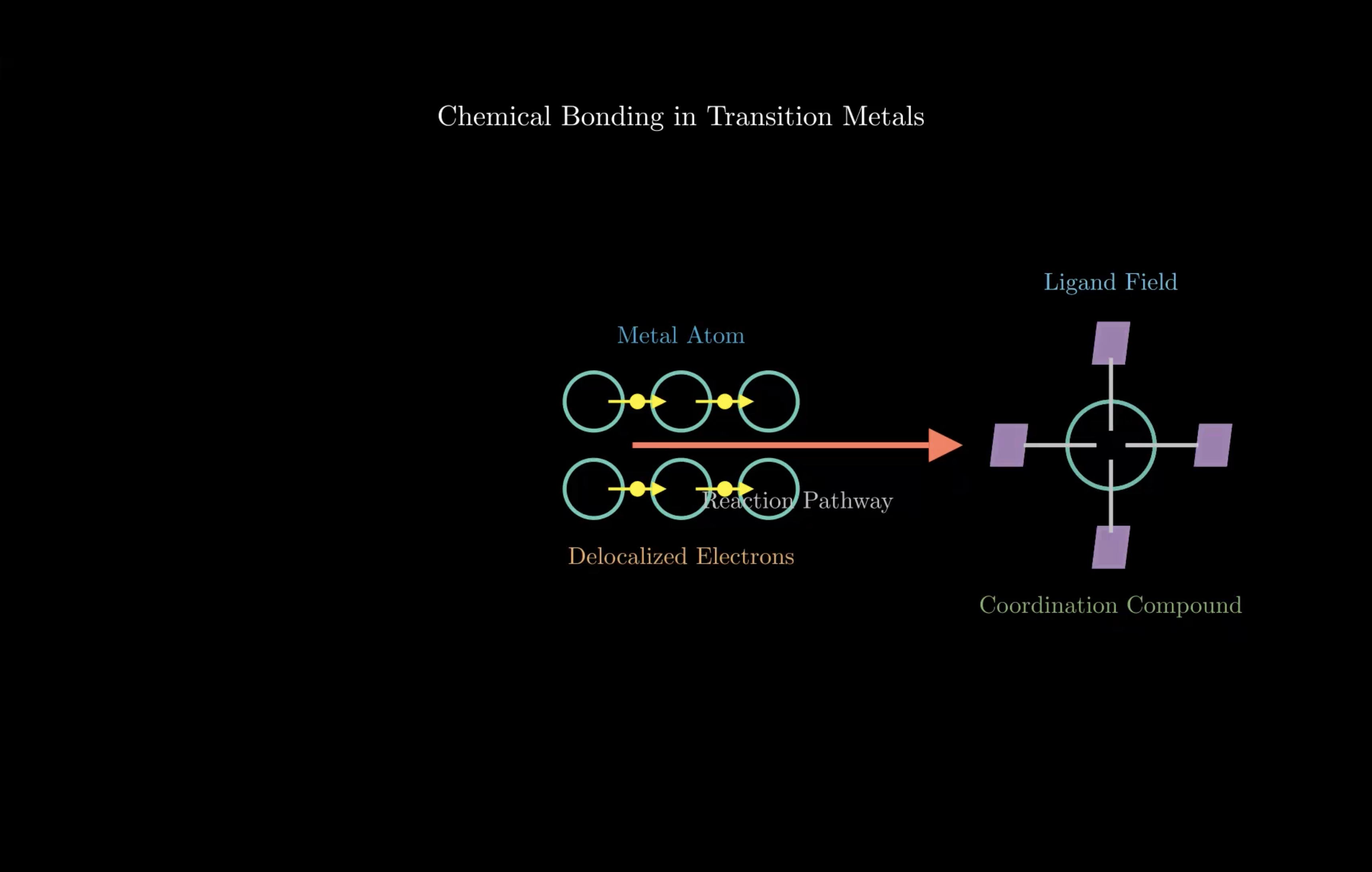}
\end{minipage}
\hfill
\begin{minipage}[t]{0.32\textwidth}
    \centering
    \includegraphics[
        width=\linewidth,
        height=0.115\textheight
    ]{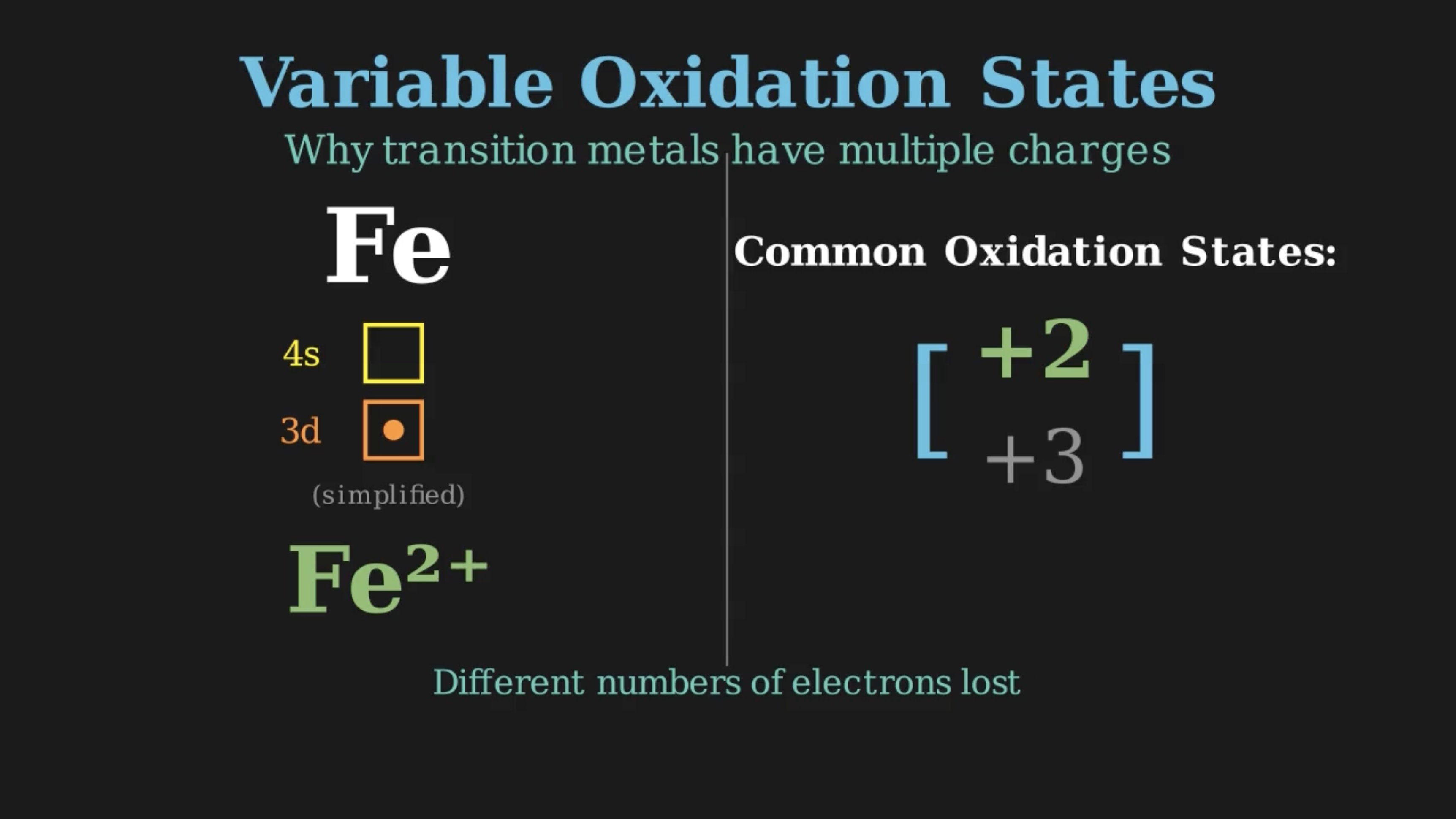}
\end{minipage}

\vspace{0.05em}

\begin{minipage}[t]{0.32\textwidth}
    \centering
    \includegraphics[
        width=\linewidth,
        height=0.115\textheight
    ]{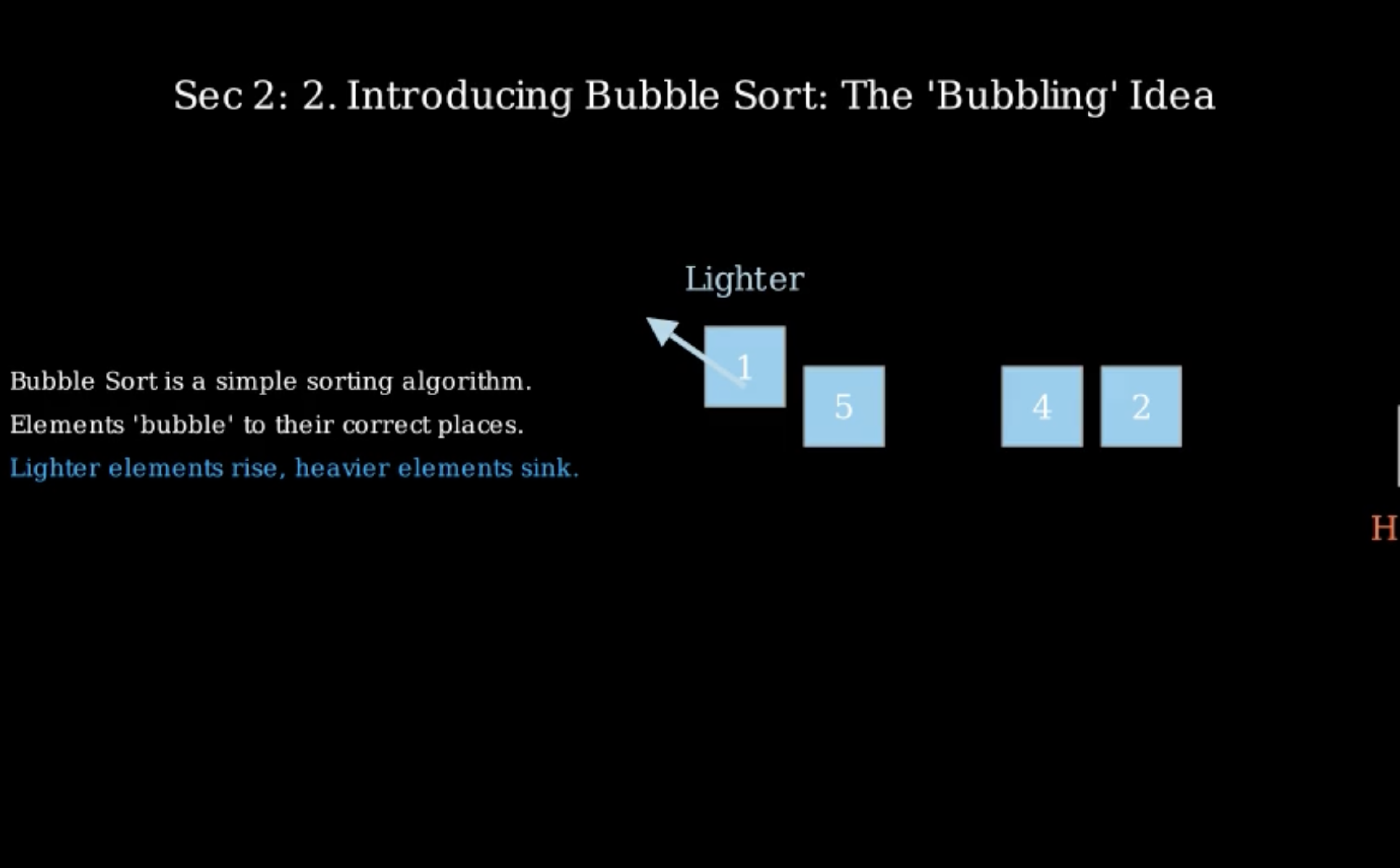}

    \vspace{0.1em}
    \scriptsize Code2Video
\end{minipage}
\hfill
\begin{minipage}[t]{0.32\textwidth}
    \centering
    \includegraphics[
        width=\linewidth,
        height=0.115\textheight
    ]{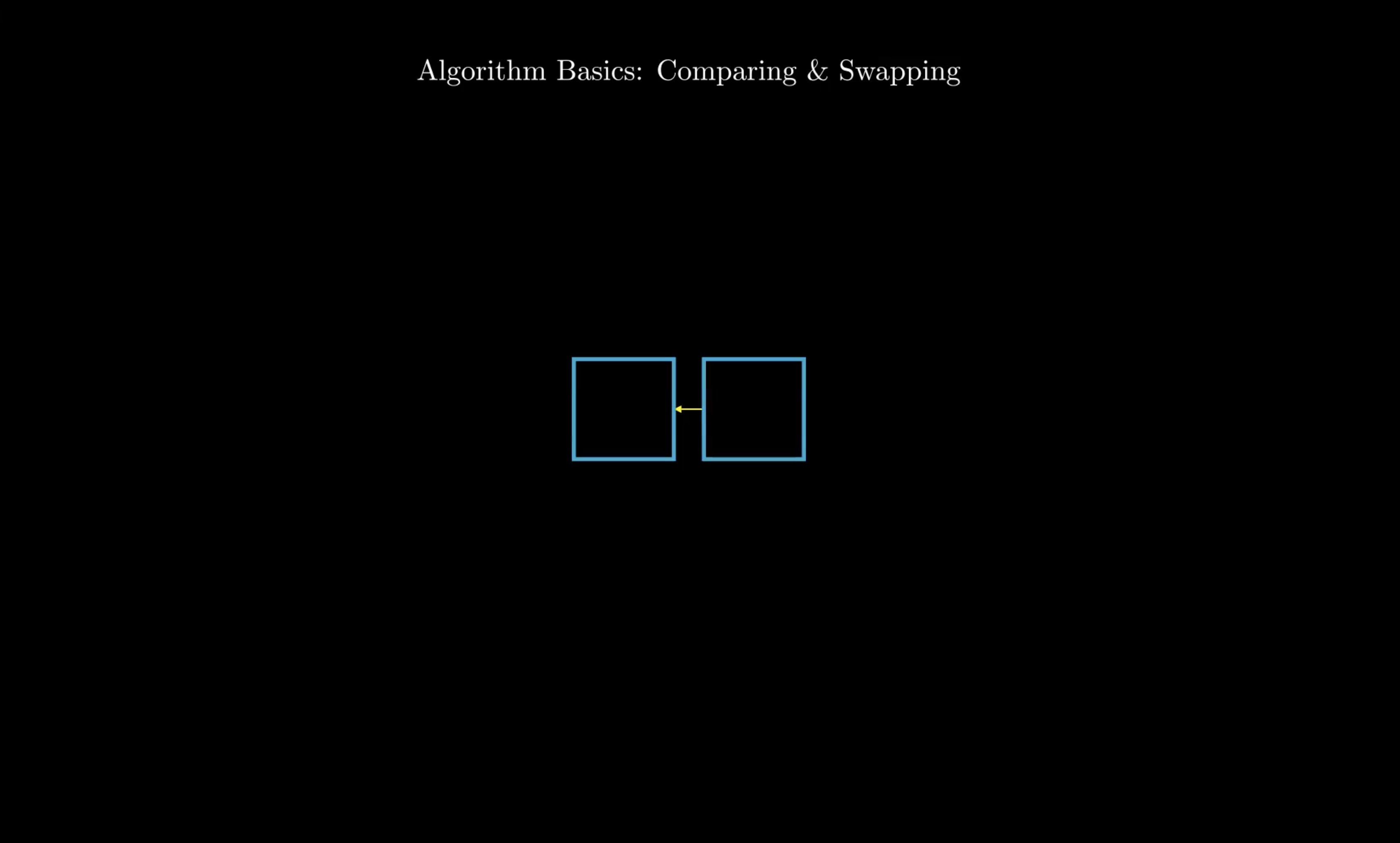}

    \vspace{0.1em}
    \scriptsize TheoremExplainAgent
\end{minipage}
\hfill
\begin{minipage}[t]{0.32\textwidth}
    \centering
    \includegraphics[
        width=\linewidth,
        height=0.115\textheight
    ]{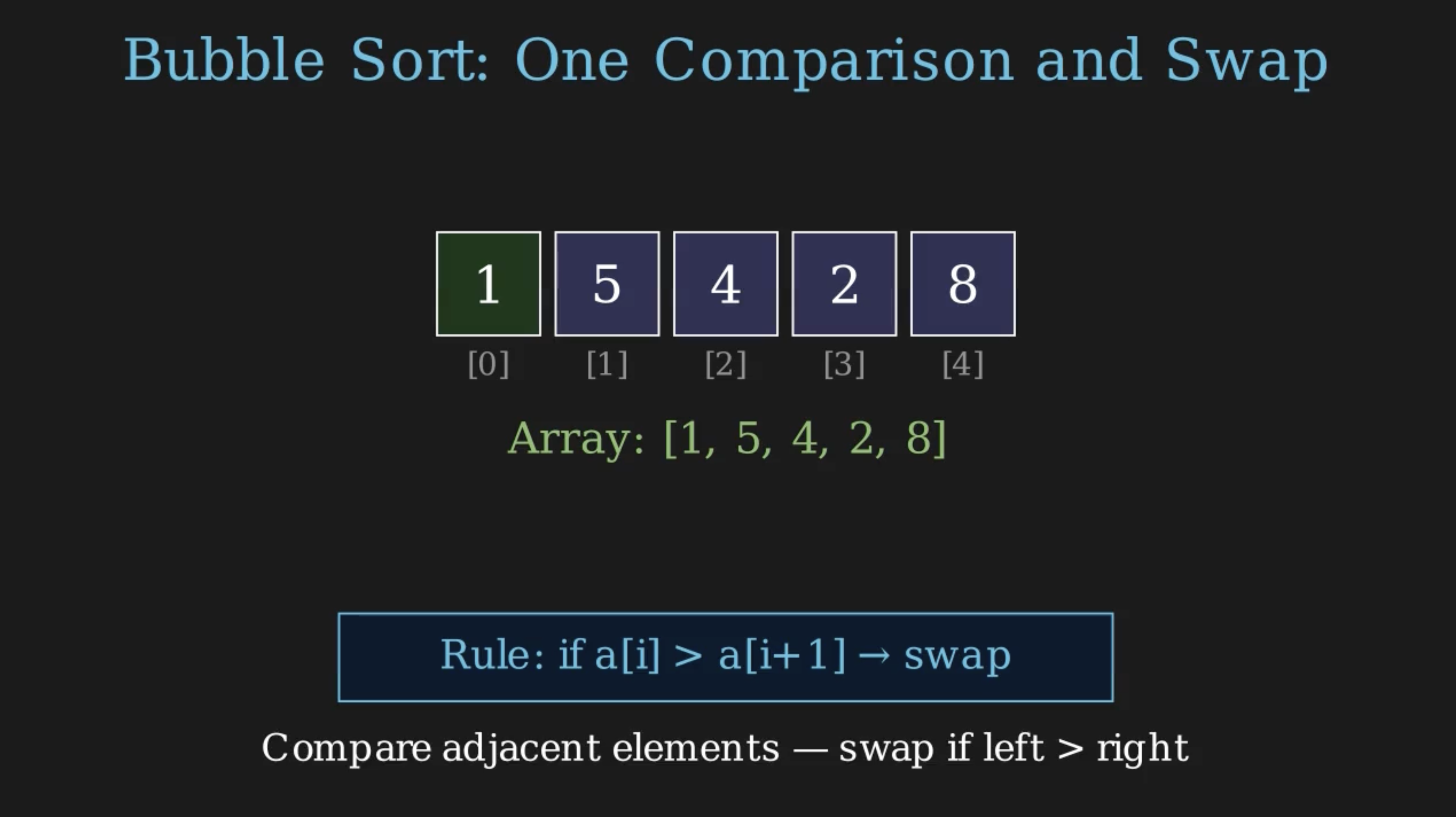}

    \vspace{0.1em}
    \scriptsize PIVOT
\end{minipage}

\vspace{-0.2em}

\caption{
Qualitative comparison of educational videos generated by different methods.
}
\label{fig:qualitative_comparison}
\end{figure*}

\subsection{Human Evaluation}


We conducted a questionnaire-based expert evaluation with 32 STEM instructors to assess the educational quality and perceived learning effects of the generated videos. The expert panel included high-school teachers in mathematics, physics, and chemistry, as well as university instructors in computer science, covering domains where abstract concepts often require careful explanation and visual representation (see Appendix ~\ref{Expert evaluation}).
Because these instructors regularly design lessons, present complex concepts, and identify student misconceptions, their judgments provide an appropriate basis for evaluating whether the generated videos are pedagogically sound and potentially useful for learning. The panel consisted of 3 physics teachers, 12 mathematics teachers, 5 chemistry teachers, and 12 computer science instructors.

We compare two instructional paradigms. Based on the previous automatic evaluation results, TheoremExplainAgent~\cite{ku2025theoremexplainagent} is selected as the strongest baseline for expert evaluation. In the baseline setting, participants only watch a single instructional video generated by TheoremExplainAgent without any subsequent assessment or corrective feedback stage.
In contrast, our framework follows a learner-centered instructional workflow. Participants first watched a pedagogically structured instructional video and reviewed a related assessment question. We then presented a plausible incorrect answer to simulate a learner misconception, followed by a remediation video generated to address that error. Participants (\ie experts) evaluated whether the remediation effectively clarified the misconception and supported conceptual understanding.

To reduce presentation bias, the display order of the baseline videos and our generated videos is randomly shuffled during the evaluation process. In total, 32 teachers and educational experts participated in the study. All questionnaire items are evaluated using a five-point Likert scale ranging from \textbf{``Strongly Disagree''} to \textbf{``Strongly Agree.''} The evaluation measures multiple instructional dimensions, including clarity, engagement, narration pace, logical flow, learning efficiency, and overall teaching effectiveness. Detailed questionnaire dimensions are provided in Appendix~\ref{Expert evaluation}.

As shown in Figure~\ref{fig:user_study_bar},
our framework consistently outperforms the baseline across nearly all evaluation dimensions. 
The improvements are particularly pronounced in Physics and Computer Science: Physics shows an average gain of $+1.67$ points over the baseline across comparable dimensions, while Computer Science improves by $+0.76$ points. In these domains, learners’ misconceptions are often tied to intermediate procedural states that are difficult to resolve through standalone instructional videos. By incorporating assessment and targeted remediation, the framework can explicitly diagnose incorrect reasoning patterns and provide more effective corrective explanations. In contrast, improvements in Mathematics and Chemistry are comparatively smaller because many evaluated topics are more symbolic and rule-driven, where procedural steps can already be conveyed relatively well through a single instructional video. 

Notably, the largest improvement is observed in \textbf{\textit{Error Explanation}}, suggesting that misconception-aware remediation improves both conceptual understanding and potential learning effectiveness by correcting learner misunderstandings. Moreover, our framework achieves higher \textbf{\textit{Clarity}} scores, indicating that pedagogy-guided instructional planning improves the readability, coherence, and interpretability of generated educational videos.
Furthermore, $96.9\%$ of participating experts reported that the workflow of ``instructional video $\rightarrow$ assessment $\rightarrow$ targeted remediation'' provides a more effective learning experience than instructional videos alone.

These results demonstrate that integrating pedagogical guidance, formative assessment, and adaptive remediation can substantially improve learner engagement and instructional effectiveness, further supporting our goal of transforming AI-generated educational videos from passive multimedia content into active learning support systems.

\subsection{Case Study}

To provide comparison across different instructional generation paradigms, Figure~\ref{fig:qualitative_comparison} presents representative educational videos generated by \textbf{Code2Video}, \textbf{TheoremExplainAgent}, and our framework on mathematics, chemistry, and computer science topics, including the Newton--Raphson Method, Transition Metals, and Bubble Sort. Compared with prior methods, our framework produces more visually organized and pedagogically structured instructional videos, with clearer symbolic visualization, more coherent instructional layouts, and richer multimodal presentation~\footnote{Additional assessment questions and bilingual case examples are provided in Appendix~\ref{Qualitative case studies}. }.

\section{Conclusion}

We presented PIVOT, a pedagogy-guided instructional video tutoring framework that reframes STEM educational video generation as a learner-centered workflow. By integrating instructional planning, pedagogically verified video generation, and assessment-based remediation, the framework connects content generation with pedagogical structure and learner feedback. 
Our evaluations show that the framework can generate high-quality educational videos that are pedagogically aligned and perceived as effective, while the integrated assessment and remediation process provides meaningful support for learner understanding and misconception correction. 
More broadly, our findings point to a human-centered direction for educational content generation: future systems should be designed not only to produce high-quality artifacts, but also to support teaching practices, learner understanding, and corrective feedback.

\section*{Limitations}
Although the proposed framework demonstrates promising capabilities in pedagogy-guided educational video generation and adaptive remediation, several limitations remain.

First, the current system primarily performs interaction-level adaptation conditioned on immediate learner responses rather than maintaining a fully personalized long-term learner model. While the framework can generate misconception-aware remediation videos for incorrect answers, it does not yet explicitly model long-term knowledge tracing, learner memory, curriculum progression, or individualized difficulty estimation across extended learning sessions.

Second, the multi-stage storyboard generation, executable rendering, verification, and remediation pipeline introduces substantially higher computational cost than direct multimedia synthesis approaches. Improving generation efficiency while preserving pedagogical quality remains an important direction for future work.

\section*{Ethics and Impact Statement}

All participants involved in the human evaluation study participated voluntarily with informed consent. The evaluation was conducted anonymously, and no personally identifiable information was collected or disclosed. Participant responses were used solely for academic research purposes and were not shared with third parties.

The study was conducted entirely online through a questionnaire-based platform, and participants could withdraw at any time without penalty. The evaluation process did not involve sensitive personal data, psychological intervention, or any procedure that could reasonably introduce ethical, physical, or social risks to participants.

More broadly, our framework is designed to support educational content creation and learner assistance in STEM education. While the system aims to improve instructional accessibility and learning support, generated educational content may still contain factual inaccuracies or pedagogical limitations. Therefore, human oversight from educators remains important when deploying such systems in real educational settings.


\section*{Acknowledgments}
The authors would like to thank the anonymous reviewers for their insightful comments. This work is supported by the National Natural Science Foundation of China (Project No. 62507015, 62537001, 62137001) and the Chenguang Program of Shanghai Education Development Foundation and Shanghai Municipal Education Commission (Project No. 24CGA26).

\bibliography{custom}

@inproceedings{ku2025theoremexplainagent,
  title={Theoremexplainagent: Towards video-based multimodal explanations for llm theorem understanding},
  author={Ku, Max and Chong, Cheuk Hei and Leung, Jonathan and Shah, Krish and Yu, Alvin and Chen, Wenhu},
  booktitle={Proceedings of the 63rd Annual Meeting of the Association for Computational Linguistics (Volume 1: Long Papers)},
  pages={6663--6684},
  year={2025}
}

@inproceedings{chen2025visualedu,
  title={VisualEDU: A Benchmark for Assessing Coding and Visual Comprehension through Educational Problem-Solving Video Generation},
  author={Chen, Hao and Shi, Tianyu and Huang, Pengran and Li, Zeyuan and Pan, Jiahui and Chen, Qianglong and He, Lewei},
  booktitle={Findings of the Association for Computational Linguistics: EMNLP 2025},
  pages={16363--16394},
  year={2025}
}

@article{liu2024sora,
  title={Sora: A review on background, technology, limitations, and opportunities of large vision models},
  author={Liu, Yixin and Zhang, Kai and Li, Yuan and Yan, Zhiling and Gao, Chujie and Chen, Ruoxi and Yuan, Zhengqing and Huang, Yue and Sun, Hanchi and Gao, Jianfeng and others},
  journal={arXiv preprint arXiv:2402.17177},
  year={2024}
}

@article{team2025kling,
  title={Kling-Omni Technical Report},
  author={Team, Kling and Chen, Jialu and Ci, Yuanzheng and Du, Xiangyu and Feng, Zipeng and Gai, Kun and Guo, Sainan and Han, Feng and He, Jingbin and He, Kang and others},
  journal={arXiv preprint arXiv:2512.16776},
  year={2025}
}

@misc{veo3_2025,
  author       = {{Google DeepMind}},
  title        = {Veo 3: Generative Video Model},
  year         = {2025},
  howpublished = {\url{https://deepmind.google/technologies/veo/}},

}

@article{mageira2022educational,
  title={Educational AI chatbots for content and language integrated learning},
  author={Mageira, Kleopatra and Pittou, Dimitra and Papasalouros, Andreas and Kotis, Konstantinos and Zangogianni, Paraskevi and Daradoumis, Athanasios},
  journal={Applied Sciences},
  volume={12},
  number={7},
  pages={3239},
  year={2022},
  publisher={MDPI}
}

@article{grassini2023shaping,
  title={Shaping the future of education: Exploring the potential and consequences of AI and ChatGPT in educational settings},
  author={Grassini, Simone},
  journal={Education sciences},
  volume={13},
  number={7},
  pages={692},
  year={2023},
  publisher={MDPI}
}

@inproceedings{lee2023lecture,
  title={Lecture presentations multimodal dataset: Towards understanding multimodality in educational videos},
  author={Lee, Dong Won and Ahuja, Chaitanya and Liang, Paul Pu and Natu, Sanika and Morency, Louis-Philippe},
  booktitle={Proceedings of the IEEE/CVF International Conference on Computer Vision},
  pages={20087--20098},
  year={2023}
}

@inproceedings{zhang2025simulating,
  title={Simulating classroom education with llm-empowered agents},
  author={Zhang, Zheyuan and Zhang-Li, Daniel and Yu, Jifan and Gong, Linlu and Zhou, Jinchang and Hao, Zhanxin and Jiang, Jianxiao and Cao, Jie and Liu, Huiqin and Liu, Zhiyuan and others},
  booktitle={Proceedings of the 2025 Conference of the Nations of the Americas Chapter of the Association for Computational Linguistics: Human Language Technologies (Volume 1: Long Papers)},
  pages={10364--10379},
  year={2025}
}

@inproceedings{chen2025code2video,
  title={Code2Video: A Code-centric Paradigm for Educational Video Generation},
  author={Chen, Yanzhe and Lin, Kevin Qinghong and Shou, Mike Zheng},
  booktitle={NeurIPS 2025 Fourth Workshop on Deep Learning for Code}
}

@article{imran2024google,
  title={Google Gemini as a next generation AI educational tool: a review of emerging educational technology},
  author={Imran, Muhammad and Almusharraf, Norah},
  journal={Smart Learning Environments},
  volume={11},
  number={1},
  pages={22},
  year={2024},
  publisher={Springer}
}

@misc{zerkouk2025comprehensivereviewaibasedintelligent,
      title={A Comprehensive Review of AI-based Intelligent Tutoring Systems: Applications and Challenges}, 
      author={Meriem Zerkouk and Miloud Mihoubi and Belkacem Chikhaoui},
      year={2025},
      eprint={2507.18882},
      archivePrefix={arXiv},
      primaryClass={cs.IR},
      url={https://arxiv.org/abs/2507.18882}, 
}

@inproceedings{pal2024autotutor,
  title={Autotutor meets large language models: A language model tutor with rich pedagogy and guardrails},
  author={Pal Chowdhury, Sankalan and Zouhar, Vil{\'e}m and Sachan, Mrinmaya},
  booktitle={Proceedings of the Eleventh ACM Conference on Learning@ Scale},
  pages={5--15},
  year={2024}
}

@inproceedings{borchers2025can,
  title={Can large language models match tutoring system adaptivity? a benchmarking study},
  author={Borchers, Conrad and Shou, Tianze},
  booktitle={International Conference on Artificial Intelligence in Education},
  pages={407--420},
  year={2025},
  organization={Springer}
}

@inproceedings{cohn2026theory,
  title={A theory of adaptive scaffolding for llm-based pedagogical agents},
  author={Cohn, Clayton and Rayala, Surya and Srivastava, Namrata and Fonteles, Joyce Horn and Jain, Shruti and Luo, Xinying and Mereddy, Divya and Mohammed, Naveeduddin and Biswas, Gautam},
  booktitle={Proceedings of the AAAI Conference on Artificial Intelligence},
  volume={40},
  number={3},
  pages={1757--1765},
  year={2026}
}

@article{hu2025generative,
  title={Generative AI in education: From foundational insights to the socratic playground for learning},
  author={Hu, Xiangen and Xu, Sheng and Tong, Richard and Graesser, Art},
  journal={arXiv preprint arXiv:2501.06682},
  year={2025}
}

@inproceedings{schmucker2024ruffle,
  title={Ruffle\&riley: Insights from designing and evaluating a large language model-based conversational tutoring system},
  author={Schmucker, Robin and Xia, Meng and Azaria, Amos and Mitchell, Tom},
  booktitle={International Conference on Artificial Intelligence in Education},
  pages={75--90},
  year={2024},
  organization={Springer}
}

@inproceedings{ye2025position,
  title={Position: LLMs Can be Good Tutors in English Education},
  author={Ye, Jingheng and Wang, Shen and Zou, Deqing and Yan, Yibo and Wang, Kun and Zheng, Hai-Tao and Liu, Ruitong and Xu, Zenglin and King, Irwin and Yu, Philip S and others},
  booktitle={Proceedings of the 2025 Conference on Empirical Methods in Natural Language Processing},
  pages={17527--17546},
  year={2025}
}

@article{graesser2004autotutor,
  title={AutoTutor: A tutor with dialogue in natural language},
  author={Graesser, Arthur C and Lu, Shulan and Jackson, George Tanner and Mitchell, Heather Hite and Ventura, Mathew and Olney, Andrew and Louwerse, Max M},
  journal={Behavior Research Methods, Instruments, \& Computers},
  volume={36},
  number={2},
  pages={180--192},
  year={2004},
  publisher={Springer}
}

@inproceedings{chen2024videocrafter2,
  title={Videocrafter2: Overcoming data limitations for high-quality video diffusion models},
  author={Chen, Haoxin and Zhang, Yong and Cun, Xiaodong and Xia, Menghan and Wang, Xintao and Weng, Chao and Shan, Ying},
  booktitle={Proceedings of the IEEE/CVF conference on computer vision and pattern recognition},
  pages={7310--7320},
  year={2024}
}

@article{zheng2025open,
  title={Open-sora 2.0: Training a commercial-level video generation model in \$200 k},
  author={Zheng, Zangwei and Peng, Xiangyu and Lou, Yuxuan and Shen, Chenhui and Young, Tom and Guo, Xinying and Wang, Binluo and Xu, Hang and Liu, Hongxin and Jiang, Mingyan and others},
  journal={arXiv preprint arXiv:2503.09642},
  year={2025}
}

@inproceedings{baek2026pedaco,
  title={PedaCo-Gen: Scaffolding Pedagogical Agency in Human-AI Collaborative Video Authoring},
  author={Baek, Injun and Kim, Yearim and Kwak, Nojun},
  booktitle={Proceedings of the Extended Abstracts of the 2026 CHI Conference on Human Factors in Computing Systems},
  pages={1--11},
  year={2026}
}

@inproceedings{dickey2024gaide,
  title={GAIDE: A framework for using generative AI to assist in course content development},
  author={Dickey, Ethan and Bejarano, Andres},
  booktitle={2024 IEEE Frontiers in Education Conference (FIE)},
  pages={1--9},
  year={2024},
  organization={IEEE}
}

@inproceedings{bar2024lumiere,
  title={Lumiere: A space-time diffusion model for video generation},
  author={Bar-Tal, Omer and Chefer, Hila and Tov, Omer and Herrmann, Charles and Paiss, Roni and Zada, Shiran and Ephrat, Ariel and Hur, Junhwa and Liu, Guanghui and Raj, Amit and others},
  booktitle={SIGGRAPH Asia 2024 Conference Papers},
  pages={1--11},
  year={2024}
}

@misc{openai2024gpt4technicalreport,
      title={GPT-4 Technical Report}, 
      author={OpenAI and Josh Achiam and Steven Adler and Sandhini Agarwal and Lama Ahmad and Ilge Akkaya and Florencia Leoni Aleman and Diogo Almeida and Janko Altenschmidt and Sam Altman and Shyamal Anadkat and Red Avila and Igor Babuschkin and Suchir Balaji and Valerie Balcom and Paul Baltescu and Haiming Bao and Mohammad Bavarian and Jeff Belgum and Irwan Bello and Jake Berdine and Gabriel Bernadett-Shapiro and Christopher Berner and Lenny Bogdonoff and Oleg Boiko and Madelaine Boyd and Anna-Luisa Brakman and Greg Brockman and Tim Brooks and Miles Brundage and Kevin Button and Trevor Cai and Rosie Campbell and Andrew Cann and Brittany Carey and Chelsea Carlson and Rory Carmichael and Brooke Chan and Che Chang and Fotis Chantzis and Derek Chen and Sully Chen and Ruby Chen and Jason Chen and Mark Chen and Ben Chess and Chester Cho and Casey Chu and Hyung Won Chung and Dave Cummings and Jeremiah Currier and Yunxing Dai and Cory Decareaux and Thomas Degry and Noah Deutsch and Damien Deville and Arka Dhar and David Dohan and Steve Dowling and Sheila Dunning and Adrien Ecoffet and Atty Eleti and Tyna Eloundou and David Farhi and Liam Fedus and Niko Felix and Simón Posada Fishman and Juston Forte and Isabella Fulford and Leo Gao and Elie Georges and Christian Gibson and Vik Goel and Tarun Gogineni and Gabriel Goh and Rapha Gontijo-Lopes and Jonathan Gordon and Morgan Grafstein and Scott Gray and Ryan Greene and Joshua Gross and Shixiang Shane Gu and Yufei Guo and Chris Hallacy and Jesse Han and Jeff Harris and Yuchen He and Mike Heaton and Johannes Heidecke and Chris Hesse and Alan Hickey and Wade Hickey and Peter Hoeschele and Brandon Houghton and Kenny Hsu and Shengli Hu and Xin Hu and Joost Huizinga and Shantanu Jain and Shawn Jain and Joanne Jang and Angela Jiang and Roger Jiang and Haozhun Jin and Denny Jin and Shino Jomoto and Billie Jonn and Heewoo Jun and Tomer Kaftan and Łukasz Kaiser and Ali Kamali and Ingmar Kanitscheider and Nitish Shirish Keskar and Tabarak Khan and Logan Kilpatrick and Jong Wook Kim and Christina Kim and Yongjik Kim and Jan Hendrik Kirchner and Jamie Kiros and Matt Knight and Daniel Kokotajlo and Łukasz Kondraciuk and Andrew Kondrich and Aris Konstantinidis and Kyle Kosic and Gretchen Krueger and Vishal Kuo and Michael Lampe and Ikai Lan and Teddy Lee and Jan Leike and Jade Leung and Daniel Levy and Chak Ming Li and Rachel Lim and Molly Lin and Stephanie Lin and Mateusz Litwin and Theresa Lopez and Ryan Lowe and Patricia Lue and Anna Makanju and Kim Malfacini and Sam Manning and Todor Markov and Yaniv Markovski and Bianca Martin and Katie Mayer and Andrew Mayne and Bob McGrew and Scott Mayer McKinney and Christine McLeavey and Paul McMillan and Jake McNeil and David Medina and Aalok Mehta and Jacob Menick and Luke Metz and Andrey Mishchenko and Pamela Mishkin and Vinnie Monaco and Evan Morikawa and Daniel Mossing and Tong Mu and Mira Murati and Oleg Murk and David Mély and Ashvin Nair and Reiichiro Nakano and Rajeev Nayak and Arvind Neelakantan and Richard Ngo and Hyeonwoo Noh and Long Ouyang and Cullen O'Keefe and Jakub Pachocki and Alex Paino and Joe Palermo and Ashley Pantuliano and Giambattista Parascandolo and Joel Parish and Emy Parparita and Alex Passos and Mikhail Pavlov and Andrew Peng and Adam Perelman and Filipe de Avila Belbute Peres and Michael Petrov and Henrique Ponde de Oliveira Pinto and Michael and Pokorny and Michelle Pokrass and Vitchyr H. Pong and Tolly Powell and Alethea Power and Boris Power and Elizabeth Proehl and Raul Puri and Alec Radford and Jack Rae and Aditya Ramesh and Cameron Raymond and Francis Real and Kendra Rimbach and Carl Ross and Bob Rotsted and Henri Roussez and Nick Ryder and Mario Saltarelli and Ted Sanders and Shibani Santurkar and Girish Sastry and Heather Schmidt and David Schnurr and John Schulman and Daniel Selsam and Kyla Sheppard and Toki Sherbakov and Jessica Shieh and Sarah Shoker and Pranav Shyam and Szymon Sidor and Eric Sigler and Maddie Simens and Jordan Sitkin and Katarina Slama and Ian Sohl and Benjamin Sokolowsky and Yang Song and Natalie Staudacher and Felipe Petroski Such and Natalie Summers and Ilya Sutskever and Jie Tang and Nikolas Tezak and Madeleine B. Thompson and Phil Tillet and Amin Tootoonchian and Elizabeth Tseng and Preston Tuggle and Nick Turley and Jerry Tworek and Juan Felipe Cerón Uribe and Andrea Vallone and Arun Vijayvergiya and Chelsea Voss and Carroll Wainwright and Justin Jay Wang and Alvin Wang and Ben Wang and Jonathan Ward and Jason Wei and CJ Weinmann and Akila Welihinda and Peter Welinder and Jiayi Weng and Lilian Weng and Matt Wiethoff and Dave Willner and Clemens Winter and Samuel Wolrich and Hannah Wong and Lauren Workman and Sherwin Wu and Jeff Wu and Michael Wu and Kai Xiao and Tao Xu and Sarah Yoo and Kevin Yu and Qiming Yuan and Wojciech Zaremba and Rowan Zellers and Chong Zhang and Marvin Zhang and Shengjia Zhao and Tianhao Zheng and Juntang Zhuang and William Zhuk and Barret Zoph},
      year={2024},
      eprint={2303.08774},
      archivePrefix={arXiv},
      primaryClass={cs.CL},
      url={https://arxiv.org/abs/2303.08774}, 
}

@book{bloom1956taxonomy,
  title={Taxonomy of educational objectives: Affective domain},
  author={Bloom, Benjamin Samuel},
  volume={2},
  year={1956},
  publisher={Longmans, Green}
}

@article{sweller1988cognitive,
  title={Cognitive load during problem solving: Effects on learning},
  author={Sweller, John},
  journal={Cognitive science},
  volume={12},
  number={2},
  pages={257--285},
  year={1988},
  publisher={Elsevier}
}

@article{renkl2014toward,
  title={Toward an instructionally oriented theory of example-based learning},
  author={Renkl, Alexander},
  journal={Cognitive science},
  volume={38},
  number={1},
  pages={1--37},
  year={2014},
  publisher={Wiley Online Library}
}

@article{mayer2005cognitive,
  title={Cognitive theory of multimedia learning},
  author={Mayer, Richard E},
  journal={The Cambridge handbook of multimedia learning},
  volume={41},
  number={1},
  pages={31--48},
  year={2005}
}

@article{black1998assessment,
  title={Assessment and classroom learning},
  author={Black, Paul and Wiliam, Dylan},
  journal={Assessment in Education: principles, policy \& practice},
  volume={5},
  number={1},
  pages={7--74},
  year={1998},
  publisher={Taylor \& Francis}
}

@article{wood1976role,
  title={The role of tutoring in problem solving},
  author={Wood, David and Bruner, Jerome S and Ross, Gail},
  journal={Journal of child psychology and psychiatry},
  volume={17},
  number={2},
  pages={89--100},
  year={1976},
  publisher={Blackwell Publishing Ltd Oxford, UK}
}

@article{alshaikh2024implementation,
  title={The implementation of the cognitive theory of multimedia learning in the design and evaluation of an AI educational video assistant utilizing large language models},
  author={AlShaikh, Rana and Al-Malki, Norah and Almasre, Maida},
  journal={Heliyon},
  volume={10},
  number={3},
  year={2024},
  publisher={Elsevier}
}

@inproceedings{calo2024towards,
  title={Towards educator-driven tutor authoring: generative AI approaches for creating intelligent tutor interfaces},
  author={Calo, Tommaso and Maclellan, Christopher},
  booktitle={Proceedings of the Eleventh ACM Conference on Learning@ Scale},
  pages={305--309},
  year={2024}
}

@inproceedings{yang2025llm,
  title={LLM-based collaborative agents with pedagogy-guided interaction modeling for timely instructive feedback generation in task-oriented group discussions},
  author={Yang, Qihao and Yang, Yu and An, Sixu and Hao, Tianyong and Xu, Guandong},
  booktitle={Proceedings of the Thirty-Fourth International Joint Conference on Artificial Intelligence},
  pages={9972--9980},
  year={2025}
}

@inproceedings{shi2025educationq,
  title={Educationq: Evaluating llms’ teaching capabilities through multi-agent dialogue framework},
  author={Shi, Yao and Liang, Rongkeng and Xu, Yong},
  booktitle={Proceedings of the 63rd Annual Meeting of the Association for Computational Linguistics (Volume 1: Long Papers)},
  pages={32799--32828},
  year={2025}
}

@article{bi2026eduillustrate,
  title={EduIllustrate: Towards Scalable Automated Generation Of Multimodal Educational Content},
  author={Bi, Shuzhen and Zhang, Mingzi and Li, Zhuoxuan and Wang, Xiaolong and Li, Keqian and Zhou, Aimin},
  journal={arXiv preprint arXiv:2604.05005},
  year={2026}
}
\clearpage

\appendix
\section{Appendix}
\label{appendix:instructional_planning_examples}

\subsection{Instructional Planning Prompt}
The prompt in Figure~\ref{fig:instructional_prompt} is used to guide the LLM to generate a pedagogically structured instructional storyboard before executable Manim code generation.

\begin{figure*}[t]
\centering

\begin{promptbox}{Instructional Planning Prompt}
\begin{Verbatim}
Given a learner-specified topic, generate a structured instructional storyboard for an educational animation video.

The output should not be executable code. Instead, it should describe the instructional design of the video. The storyboard must organize the topic into pedagogically meaningful components, including learning objectives, prerequisite activation, concept explanation, visual representation plan, worked examples, and diagnostic assessment probes.

For each topic, produce a JSON object with the following structure:


{
  "title": "...",
  "topic": "...",
  "difficulty_level": "...",
  "learning_objective": "...",
  "prerequisite_activation": "...",
  "concept_explanations": [
    {
      "concept": "...",
      "explanation_goal": "...",
      "scaffolding_strategy": "..."
    }
  ],
  "visual_plan": "...",
  "worked_example": "...",
  "assessment_probe": "...",
  "shots": [
    {
      "shot": 1,
      "goal": "...",
      "narration": "...",
      "visual_description": "...",
      "visual_elements": ["...", "..."],
      "duration": "..."
    }
  ]
}


Design requirements:
1. Each shot should focus on one local learning goal.
2. The storyboard should progress from prerequisite knowledge to core concepts, then to examples and assessment.
3. Visual plans should specify formulas, diagrams, highlights, arrows, labels, and animations where appropriate.
4. Narration should explain why the concept works, not only describe what appears on screen.
5. The assessment probe should target a likely misconception or reasoning error.
6. Output only valid JSON.
\end{Verbatim}
\end{promptbox}

\caption{Prompt template used for instructional planning.}
\label{fig:instructional_prompt}
\end{figure*}

\subsection{Generated Storyboard Example}
For the learner-specified topic “Bubble Sort”, the instructional planning module generates the corresponding pedagogical storyboard, as illustrated in Figure~\ref{fig:Pedagogical Storyboard of Bubble Sort}.

\begin{figure*}[t]
\centering
\begin{examplebox}{Generated Storyboard Example}
\begin{Verbatim}
{
  "title": "Bubble Sort: Moving the Largest Element Step by Step",
  "topic": "Bubble Sort",
  "difficulty_level": "Basic",
  "learning_objective": "Learners will understand how bubble sort repeatedly compares adjacent elements, swaps inverted pairs, and places the largest remaining element at the end of the unsorted region after each pass.",
  "prerequisite_activation": "Recall the concepts of arrays, element indices, comparison operators, loops, and swapping two values.",
  "concept_explanations": [
    {
      "concept": "Adjacent comparison",
      "explanation_goal": "Show that bubble sort only compares neighboring elements rather than searching the whole array at once.",
      "scaffolding_strategy": "Begin with one pair of elements, display the comparison condition, and animate either swap or no-swap."
    },
    {
      "concept": "One full pass",
      "explanation_goal": "Explain why the largest value must move to the rightmost unsorted position after one pass.",
      "scaffolding_strategy": "Track the same large element across successive adjacent comparisons until it reaches the end."
    },
    {
      "concept": "Shrinking unsorted region",
      "explanation_goal": "Show how each completed pass reduces the range that still needs to be sorted.",
      "scaffolding_strategy": "Use a locked or greyed-out region to mark elements that no longer participate in later passes."
    }
  ],
  "visual_plan": "Represent the array as a row of numbered rectangles. Use yellow outlines to highlight the current adjacent pair, red arrows for swaps, green locks for elements that are fixed, and a grey overlay for the sorted suffix. Display loop variables i and j near the array to connect the animation with the algorithm.",
  "worked_example": "Use the array [5, 1, 4, 2, 8]. Demonstrate the first pass: compare 5 and 1, swap; compare 5 and 4, swap; compare 5 and 2, swap; compare 5 and 8, no swap. Conclude that 8 is fixed at the end.",
  "assessment_probe": "After the first pass of bubble sort on [3, 1, 4, 2], which element is guaranteed to be in its final position? Explain why."
}
\end{Verbatim}
\end{examplebox}
\caption{Pedagogical Storyboard of Bubble Sort.}
\label{fig:Pedagogical Storyboard of Bubble Sort}
\end{figure*}

\subsection{Shot-level Storyboard Example}
As shown in Figure~\ref{fig:Shot-Level Storyboard of Bubble Sort} , the high-level instructional plan is further decomposed into shot-level storyboard units. Each shot contains a local learning goal, narration, visual specification, expected duration, and visual elements that guide downstream Manim generation.

\begin{figure*}[t]
\centering

\begin{examplebox}{Shot-level Storyboard Example}
\begin{Verbatim}
[
  {
    "shot": 1,
    "goal": "Introduce the intuition of bubble sort.",
    "narration": "Bubble sort works by repeatedly comparing two neighboring values. If they are in the wrong order, we swap them. After enough adjacent swaps, larger values gradually move toward the right side of the array.",
    "visual_description": "Show the title 'Bubble Sort' and an array [5, 1, 4, 2, 8]. A small moving marker travels from left to right above adjacent pairs, indicating the direction of comparison.",
    "visual_elements": [
      "title text",
      "array rectangles",
      "comparison arrow",
      "moving marker",
      "rightward direction cue"
    ],
    "duration": "12 seconds"
  },
...
  {
    "shot": 4,
    "goal": "Connect the animation to nested loops.",
    "narration": "The outer loop counts how many passes have been completed. The inner loop performs adjacent comparisons only inside the remaining unsorted region.",
    "visual_description": "Place pseudocode on the left and the animated array on the right. Highlight loop variable i in blue and j in orange. Grey out the sorted suffix.",
    "visual_elements": [
      "pseudocode block",
      "i highlight",
      "j pointer",
      "grey sorted suffix",
      "range bracket"
    ],
    "duration": "12 seconds"
  },
...
]
\end{Verbatim}
\end{examplebox}

\caption{Pedagogical Shot-Level Storyboard of Bubble Sort.}
\label{fig:Shot-Level Storyboard of Bubble Sort}
\end{figure*}

\subsection{Constrained Layout Prompt}
\label{appendix:Constrained Layout Prompt}
Figure~\ref{fig:Layout Prompt} presents the constrained layout prompt used to improve spatial organization, visual readability, and instructional consistency during executable educational video generation.

\begin{figure*}[t]
\centering
\begin{promptbox}{Constrained Layout Prompt}
\begin{Verbatim}
You are a Manim animation programming expert and visual designer. Your task is to generate beautiful and professional Manim animation code based on an educational video storyboard.
[Most Important] Visual Layout Quality Requirements (must be strictly followed):
1. Screen Layout
   - Place the title near the top of the screen, around UP * 3.
   - Keep the main content centered.
   - Place explanatory text near the bottom of the screen, around DOWN * 3.
   - Leave sufficient margins on the left and right; do not place elements too close to the edges.
2. Object Spacing
   - Maintain sufficient spacing between elements, using buff=0.5 or larger.
   - Do not allow formulas, diagrams, labels, or text to overlap.
   - Use .arrange() to automatically align and space multiple elements.
3. Font Size Hierarchy
   - Main title: font_size=56
   - Subtitle: font_size=40
   - Body text: font_size=32
   - Notes or labels: font_size=24
...
[Key Requirement] Animation Sequencing Control to Prevent Overlap:
- Before introducing new content, first clear the previous content with `FadeOut`.
- Alternatively, use `Transform` or `ReplacementTransform` to replace old content.
- Do not stack multiple unrelated elements on top of each other.
- Very important: Do not use `FadeOut` to clear the screen at the end of a shot.
...
Layout Checklist:
- [ ] Are elements centered or reasonably aligned?
- [ ] Is there sufficient spacing between elements?
- [ ] Is the font size hierarchy clear?
- [ ] Are the colors visually coordinated?
- [ ] Are there any overlapping elements?
\end{Verbatim}
\end{promptbox}
\caption{Constrained Layout Prompt for Instructional Video Generation.}
\label{fig:Layout Prompt}
\end{figure*}

\section{Evaluation}

\subsection{Instructional Topics Used for Video Generation and Evaluation}
\label{Instructional Topics Used for Video Generation and Evaluation}
The knowledge points in Figure~\ref{fig:Knowledge Points} were used for instructional video generation and evaluation in our experiments.

\begin{figure*}[t]
\centering
\begin{goldbox}{Knowledge Points}
\begin{Verbatim}[commandchars=\\\{\}]
\textbf{Mathematics}
Fermat's Theorem, Newton--Raphson Method, Arithmetic Sequence, Congruent Triangle Theorem, Eccentricity, Euler's Formula, Geometric Sequence, Heron's Formula, Inverse Function, Laws of Exponents

\textbf{Computer Science}
Bubble Sort, Deadlock, Huffman Coding, IEEE Floating-Point Representation, ALOHA Protocol, Properties of Binary Trees, Pigeonhole Principle, Karnaugh Map, Run-Length Encoding, Signal-to-Noise Ratio

\textbf{Chemistry}
Transition Metals, Chemical Equation Balancing, Combustion Analysis, Crystallization, Distillation, First Law of Thermodynamics, Law of Conservation of Mass, Noble Gases, Oxidation-Reduction Reactions, Titration

\textbf{Physics}
Avogadro Constant, Color and Wavelength, Coulomb's Law, Dalton's Law of Partial Pressures, Universal Gravitation, Kinetic Energy, Newton's First Law of Motion, Newton's Second Law of Motion, Ohm's Law, Archimedes' Principle
\end{Verbatim}
\end{goldbox}
\caption{Knowledge Points in Experiment.}
\label{fig:Knowledge Points}
\end{figure*}

\subsection{Automatic Evaluation}
\label{Automatic Evaluation}

Figure~\ref{fig:Visual Evaluation Prompt} and Figure~\ref{fig:Narration Evaluation Prompt} present the prompts used for automatic evaluation of generated instructional videos and narrations. 

\begin{figure*}[t]
\centering
\begin{promptbox}{Transcript Evaluation Prompt}
\begin{Verbatim}[commandchars=\\\{\}]
You are a specialist in evaluating theorem explanation videos, known for giving clear and objective feedback. You will be given the transcript of a video. Your task is to evaluate and score the content of the video in several dimensions.
\textbf{Evaluation Criteria:}
\textbf{1. Accuracy and Depth}
• Does the narration explain the theorem accurately?
• Does the video provide intuitive and/or rigorous explanations for why the theorem holds?
\textbf{2. Logical Flow}
• Does the video follow a clear and logical structure?
• Does the video present a coherent buildup of ideas?


\end{Verbatim}
\end{promptbox}
\caption{Narration Evaluation Prompt.}
\label{fig:Narration Evaluation Prompt}
\end{figure*}

\begin{figure*}[t]
\centering
\begin{promptbox}{Visual Layout Evaluation Prompt}
\begin{Verbatim}
Please evaluate the image in terms of Spatial Clarity and Layout quality. 
Assign a score from 0 to 10 based on the following dimensions:

Evaluation Dimensions:
1. Visual Cleanliness (40%)
   - Is the background clean and free of distracting or cluttered elements?

2. Occlusion and Overlap (30%)
   - Do text, diagrams, or graphical elements overlap or block each other?

3. Layout Comfort and Readability (30%)
   - Is the layout well organized?
   - Is whitespace used appropriately?
   - Is the visual flow clear and easy to follow?
...
\end{Verbatim}
\end{promptbox}
\caption{Visual Evaluation Prompt.}
\label{fig:Visual Evaluation Prompt}
\end{figure*}







\subsection{Expert evaluation}
\label{Expert evaluation}
The expert evaluation was conducted fully online using a questionnaire-based assessment platform. We randomly recruited 32 STEM instructors from public educational discussion forums and teacher communities, including high-school teachers in mathematics, physics, and chemistry, as well as university instructors in computer science. This recruitment strategy helps improve diversity in teaching backgrounds and reduces potential sampling bias associated with a single institution or local region, providing a reasonably representative sample for the user study.

\begin{figure*}[t]
    \centering

    \begin{subfigure}[t]{0.48\linewidth}
        \centering
        \includegraphics[
            width=\linewidth,
            height=0.35\textheight
        ]{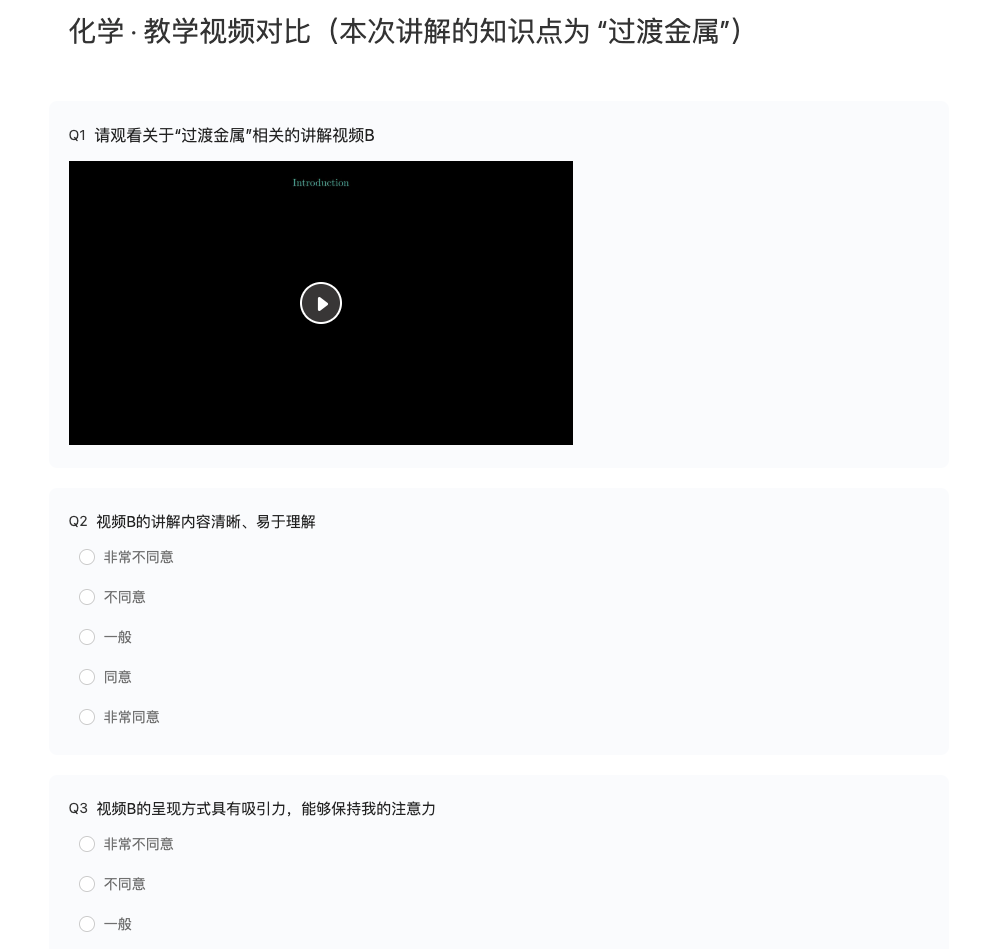}
        \caption{Chinese questionnaire}
    \end{subfigure}
    \hfill
    \begin{subfigure}[t]{0.48\linewidth}
        \centering
        \includegraphics[
            width=\linewidth,
            height=0.35\textheight
        ]{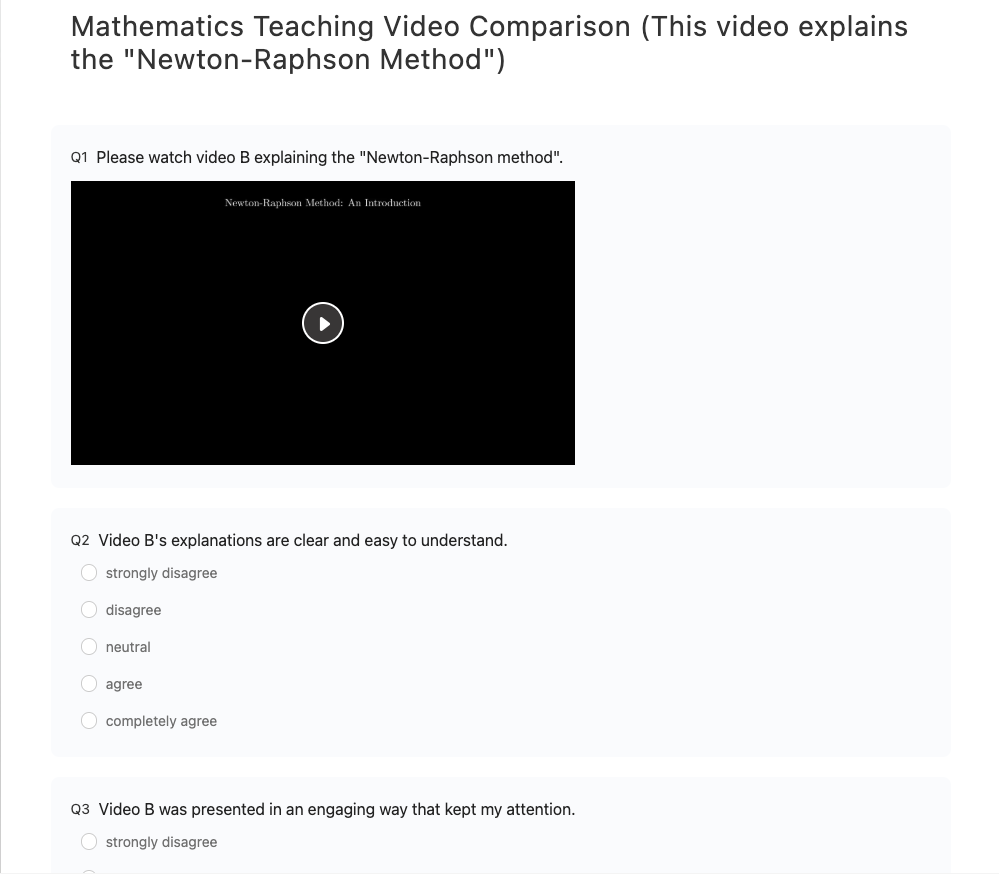}
        \caption{English questionnaire}
    \end{subfigure}

    \caption{Online questionnaire interface used in the expert evaluation study (Chinese and English versions).}
    \label{fig:questionnaire_interface}
\end{figure*}

All participants independently evaluated the generated instructional videos without interaction with the authors or other participants. The questionnaire included multiple pedagogically grounded evaluation dimensions, including instructional clarity, engagement, narration pace, logical flow, learning efficiency, and overall teaching effectiveness. Table~\ref{tab:metric_mapping} summarizes the abbreviations of questionnaire items used in the evaluation. To improve transparency and reproducibility, screenshots of the online evaluation interface and questionnaire examples are provided in Figure~\ref{fig:questionnaire_interface}.

\begin{table*}[!t]
\centering
\small
\setlength{\tabcolsep}{8pt}
\renewcommand{\arraystretch}{1.15}

\begin{tabular}{p{0.22\linewidth} p{0.68\linewidth}}
\toprule
\textbf{Metric} & \textbf{Questionnaire Item} \\
\midrule

Clarity &
The explanation content is clear and easy to understand.
\\

Engagement &
The presentation style is engaging and keeps my attention.
\\

Narration Pace &
The narration pace is appropriate.
\\

Logical Flow &
The instructional structure and logical flow are clear.
\\

Error Explanation &
The error-explanation video clearly explains each answer option.
\\

Learning Efficiency &
This teaching method improves my learning efficiency.
\\

Teaching Effectiveness &
This teaching method improves overall teaching effectiveness.
\\

\bottomrule
\end{tabular}

\caption{Mapping between evaluation metrics and questionnaire items used in the human evaluation study.}
\label{tab:metric_mapping}

\vspace{-0.5em}
\end{table*}











\section{Qualitative case studies }
\label{Qualitative case studies}

\subsection{Examples of Assessment Questions }
\label{Examples of Assessment Questions}
As shown in Figure~\ref{fig:assessment questions}, the generated post-instruction assessment questions for transition metals and gradient descent are designed to identify learners’ potential misconceptions and reasoning errors.

\begin{figure*}[t]
\centering

\begin{modelbox}{}
\begin{Verbatim}[commandchars=\\\{\}]
\textbf{Which of the following statements about transition metals is incorrect?}


\textbf{A.} Transition metals are mainly located in the \textit{d}-block of the periodic table, corresponding to Groups 3--12.
    
\textbf{B.} Transition metals often exhibit multiple oxidation states. For example, manganese can exist in oxidation states such as +2, +3, +4, +6, and +7.
    
\textbf{C.} The colors of transition metal compounds are related to electron transitions among partially filled \textit{d} orbitals.
    
\textbf{D.} The valence electrons of transition metals come only from the outermost \textit{ns} orbital, while the \((n-1)d\) orbitals do not participate in chemical reactions.
...
\textbf{You are minimizing a differentiable loss function and you are currently at a point where the slope (gradient) is positive. To reduce the loss using gradient descent, which update behavior is most appropriate, and what would you expect if the learning rate is set far too large?}

\textbf{A.} Move in the negative gradient direction; an overly large learning rate can overshoot the minimum and cause oscillation or divergence.
    
\textbf{B.} Move in the positive gradient direction; an overly large learning rate will converge faster without risk.
    
\textbf{C.} Do not move because a positive gradient indicates you are already at the minimum; an overly large learning rate has no effect.
    
\textbf{D.} Move in a random direction to escape local minima; an overly large learning rate mainly slows progress but remains stable.
\end{Verbatim}
\end{modelbox}

\caption{Post-instruction assessment questions generated from instructional videos on transition metals and gradient descent.}
\label{fig:assessment questions}
\end{figure*}


    
    
    


\subsection{Bilingual Case Analysis }
\label{Bilingual Case Analysis}

The generated workflow includes pedagogically structured instructional videos together with assessment-driven remediation videos. The framework supports controllable multimodal instructional generation, including symbolic visualization, high-fidelity diagrams, and dynamic instructional animations. Moreover, the generated remediation videos remain aligned with the corresponding assessment questions and learner misunderstanding patterns, enabling misconception-aware corrective explanations rather than generic answer summaries.

\begin{figure*}[t]
\centering

\begin{minipage}[t]{0.48\textwidth}
    \centering
    \includegraphics[
        width=\linewidth,
        height=0.18\textheight
    ]{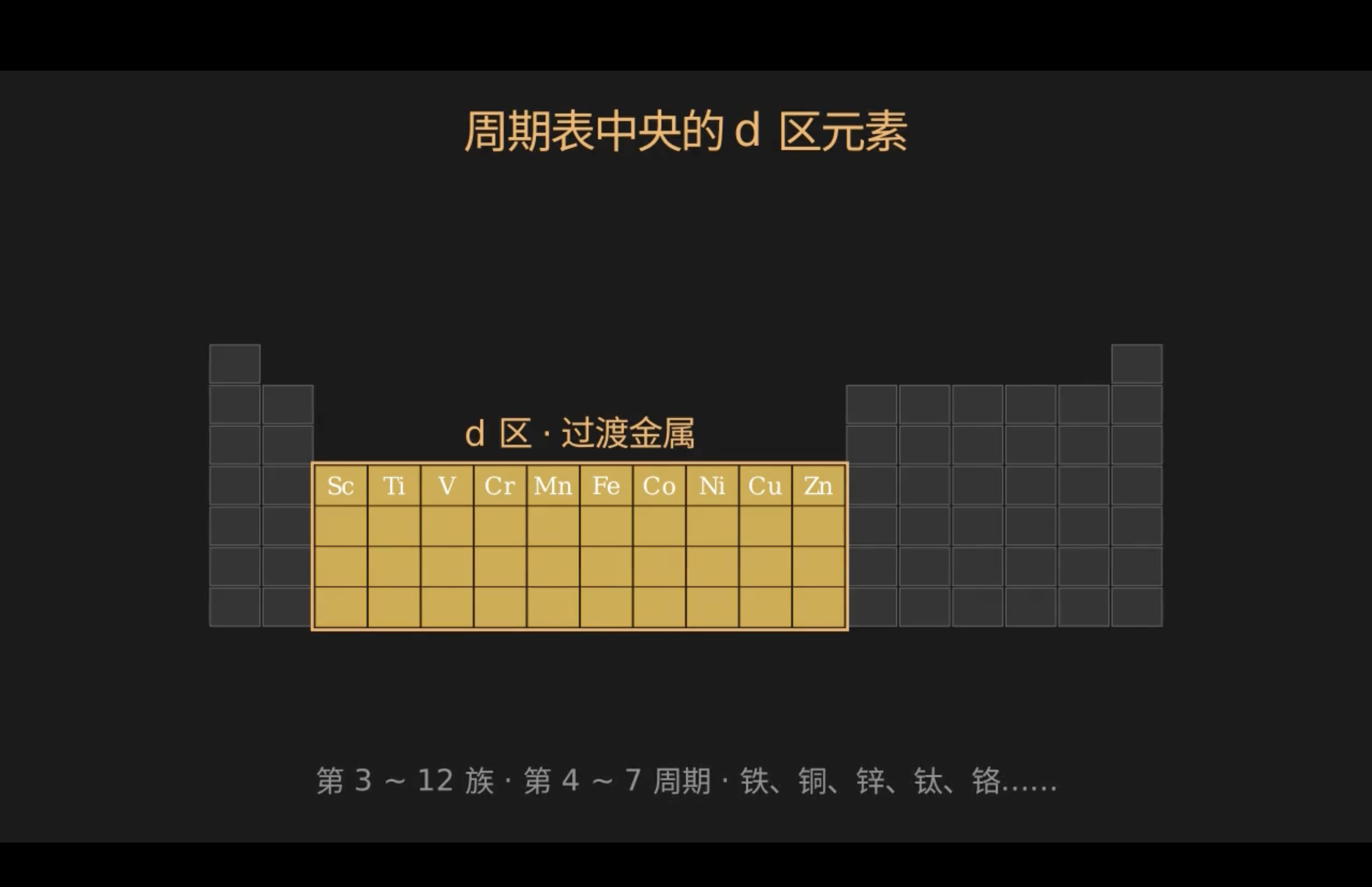}

    \vspace{0.2em}
    \small (a) Chinese chemistry instructional video
\end{minipage}
\hfill
\begin{minipage}[t]{0.48\textwidth}
    \centering
    \includegraphics[
        width=\linewidth,
        height=0.18\textheight
    ]{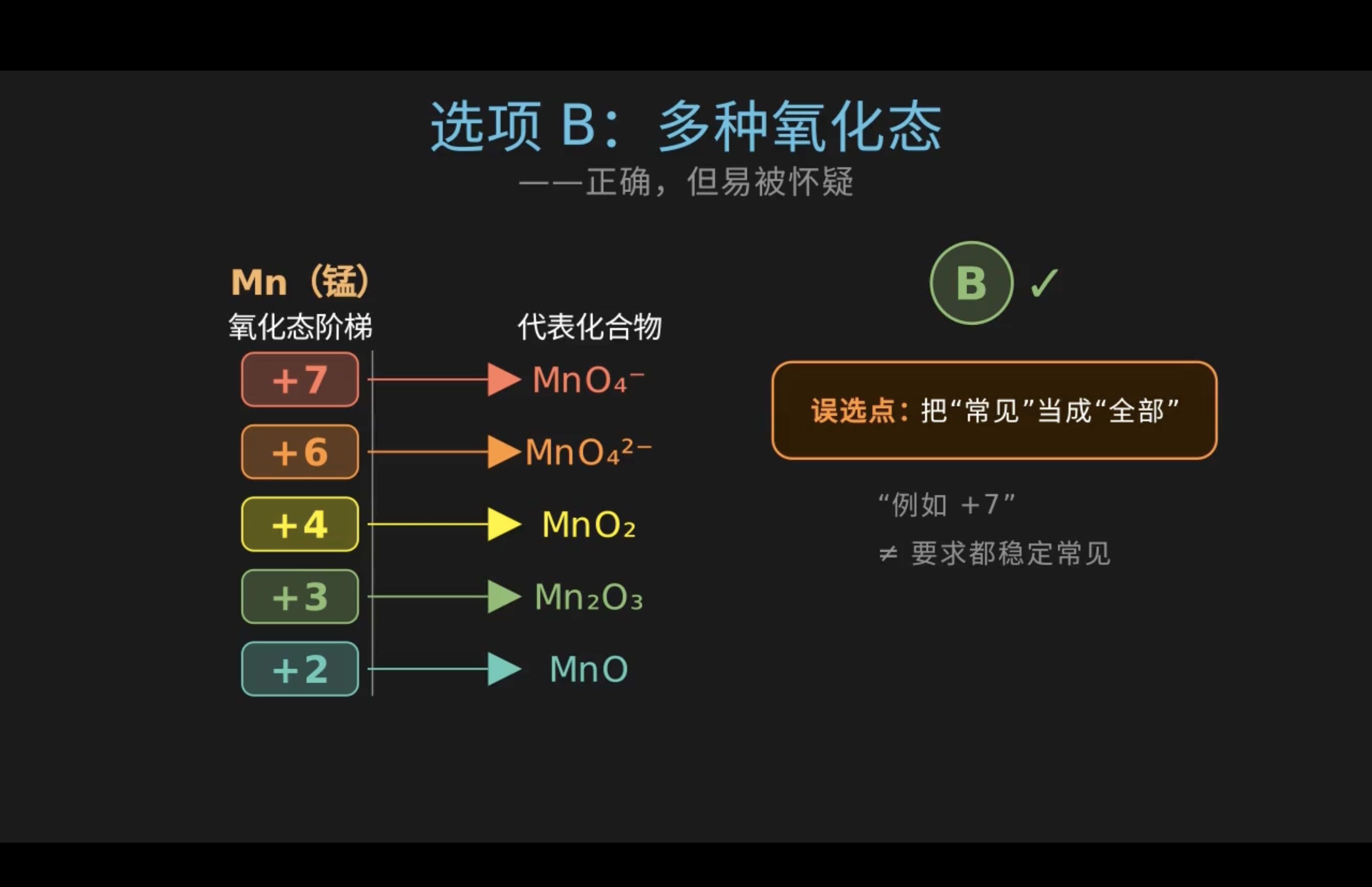}

    \vspace{0.2em}
    \small (b) Chinese chemistry remediation video
\end{minipage}

\vspace{0.8em}

\begin{minipage}[t]{0.48\textwidth}
    \centering
    \includegraphics[
        width=\linewidth,
        height=0.18\textheight
    ]{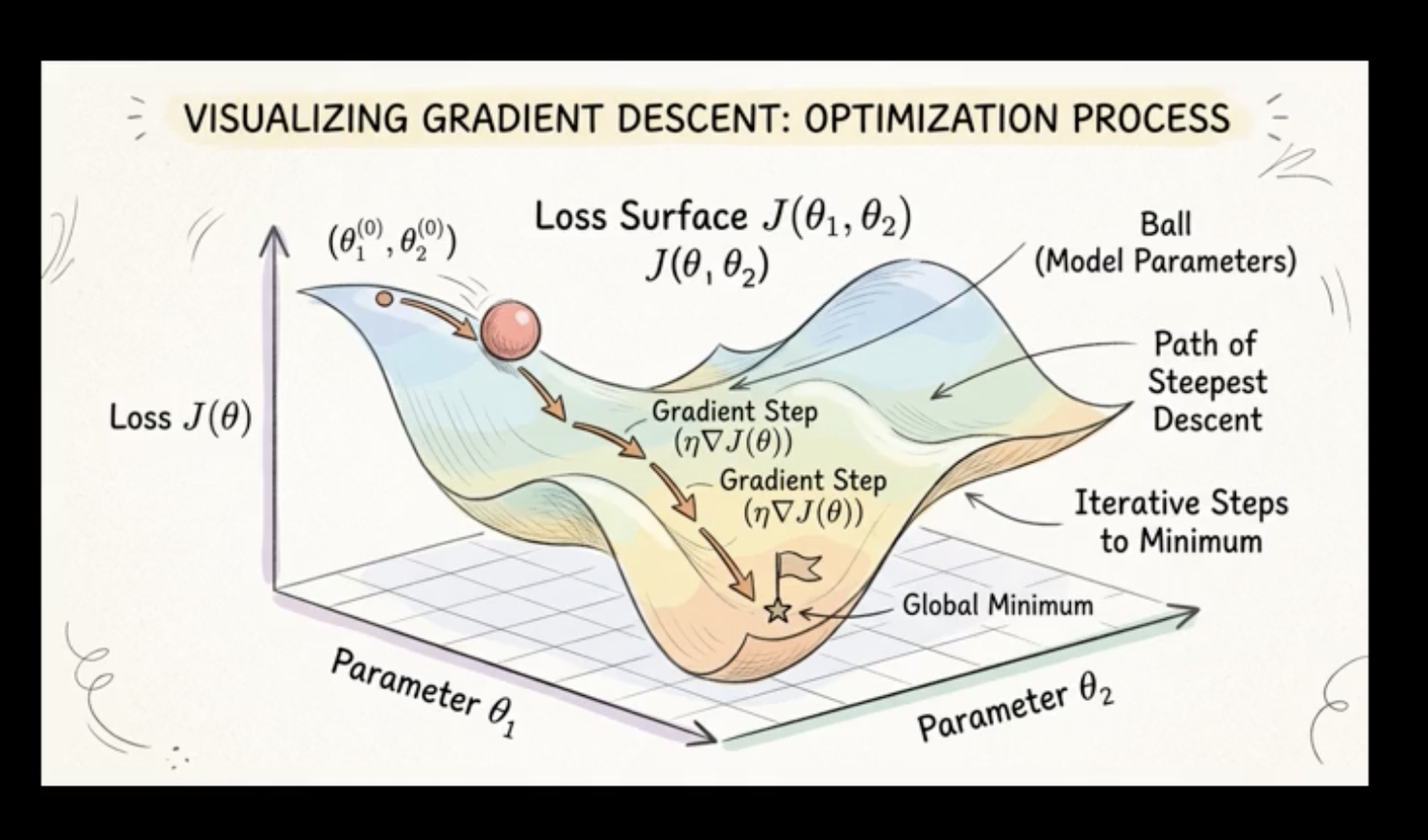}

    \vspace{0.2em}
    \small (c) English gradient descent instructional video
\end{minipage}
\hfill
\begin{minipage}[t]{0.48\textwidth}
    \centering
    \includegraphics[
        width=\linewidth,
        height=0.18\textheight
    ]{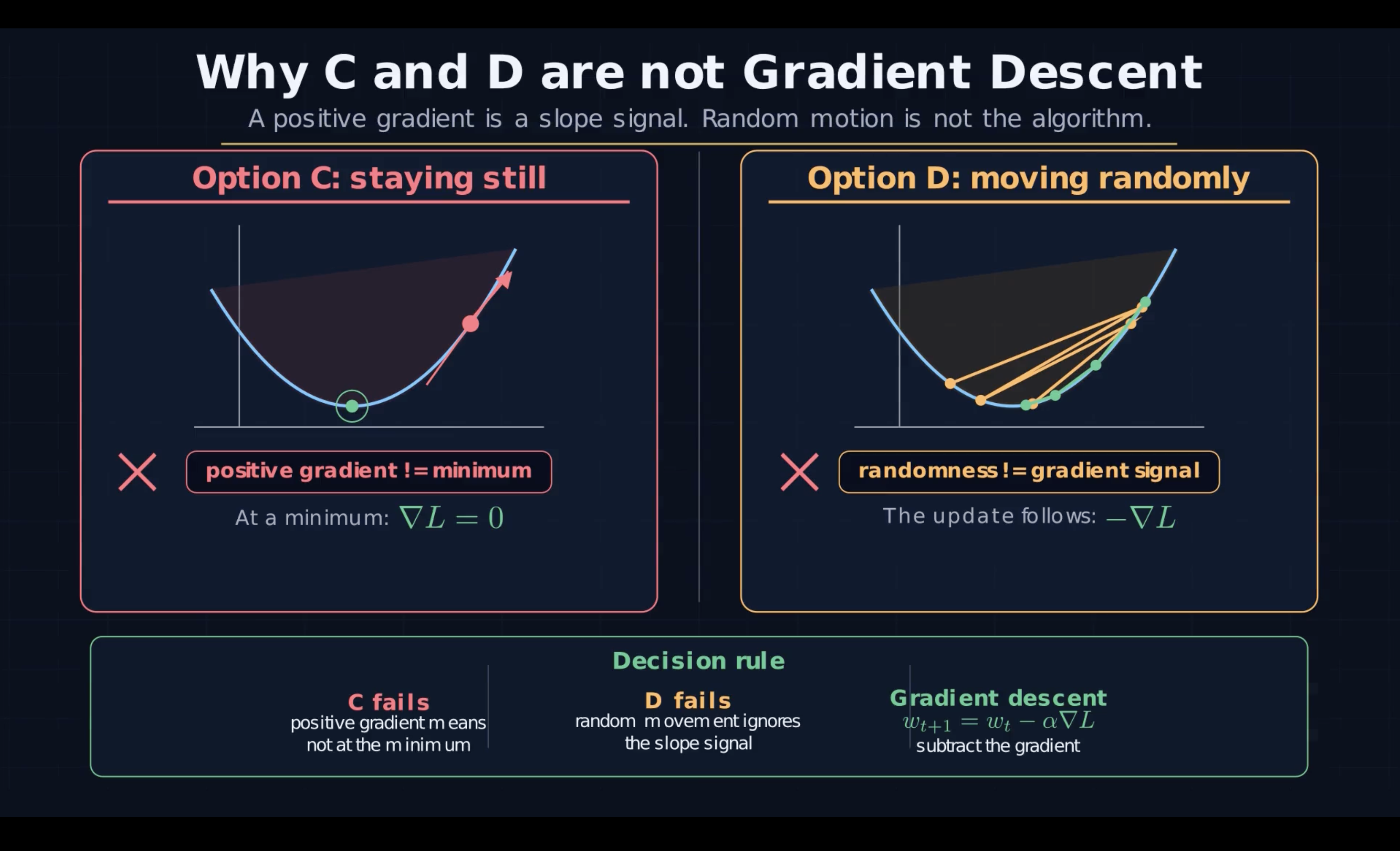}

    \vspace{0.2em}
    \small (d) English gradient descent remediation video
\end{minipage}

\caption{
Qualitative case studies generated under bilingual educational settings.
}
\label{fig:bilingual_case_study}
\end{figure*}

\begin{figure*}[t]
\centering

\begin{minipage}[t]{0.48\textwidth}
    \centering
    \includegraphics[
        width=\linewidth,
        height=0.18\textheight
    ]{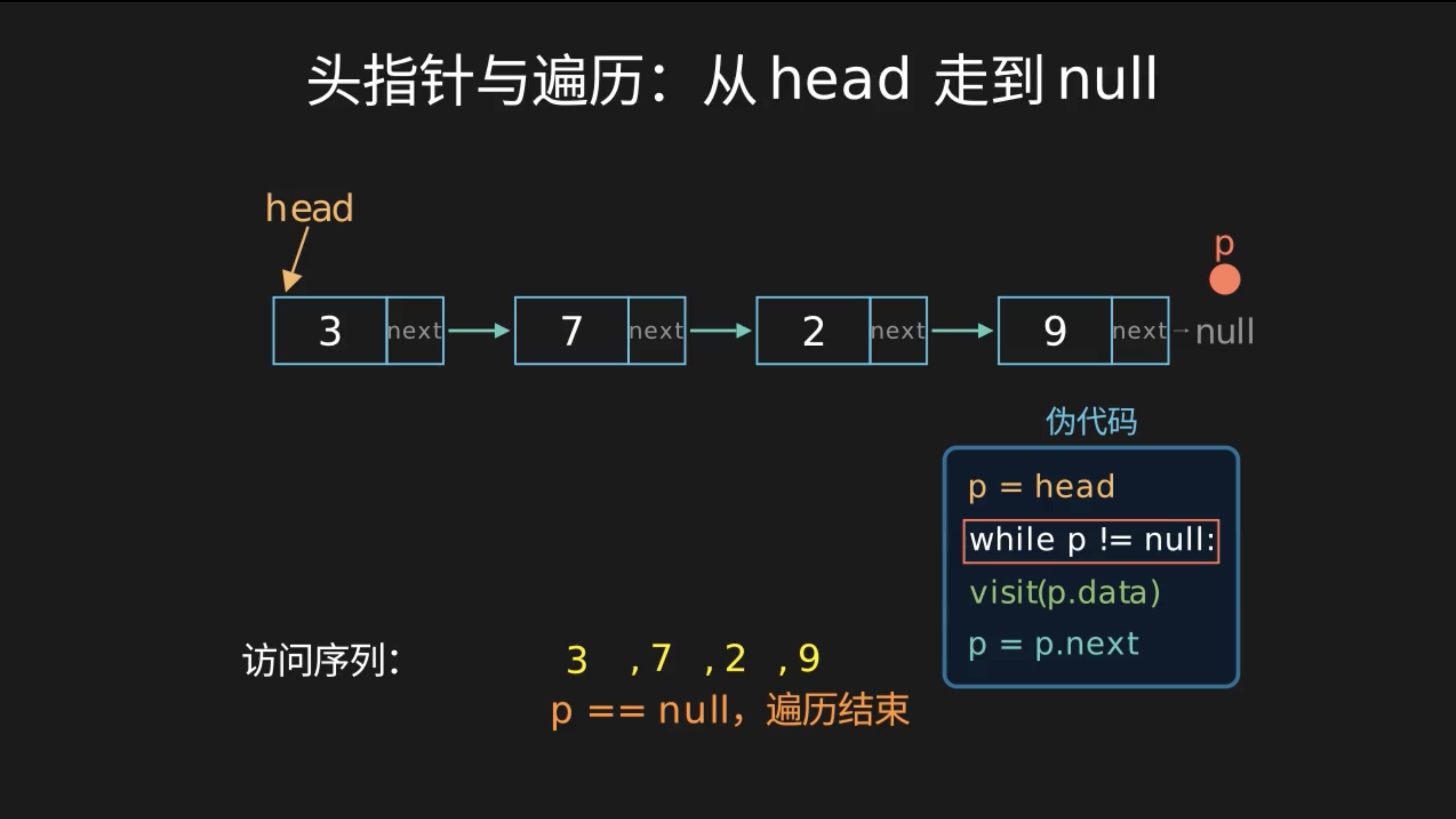}
\end{minipage}
\hfill
\begin{minipage}[t]{0.48\textwidth}
    \centering
    \includegraphics[
        width=\linewidth,
        height=0.18\textheight
    ]{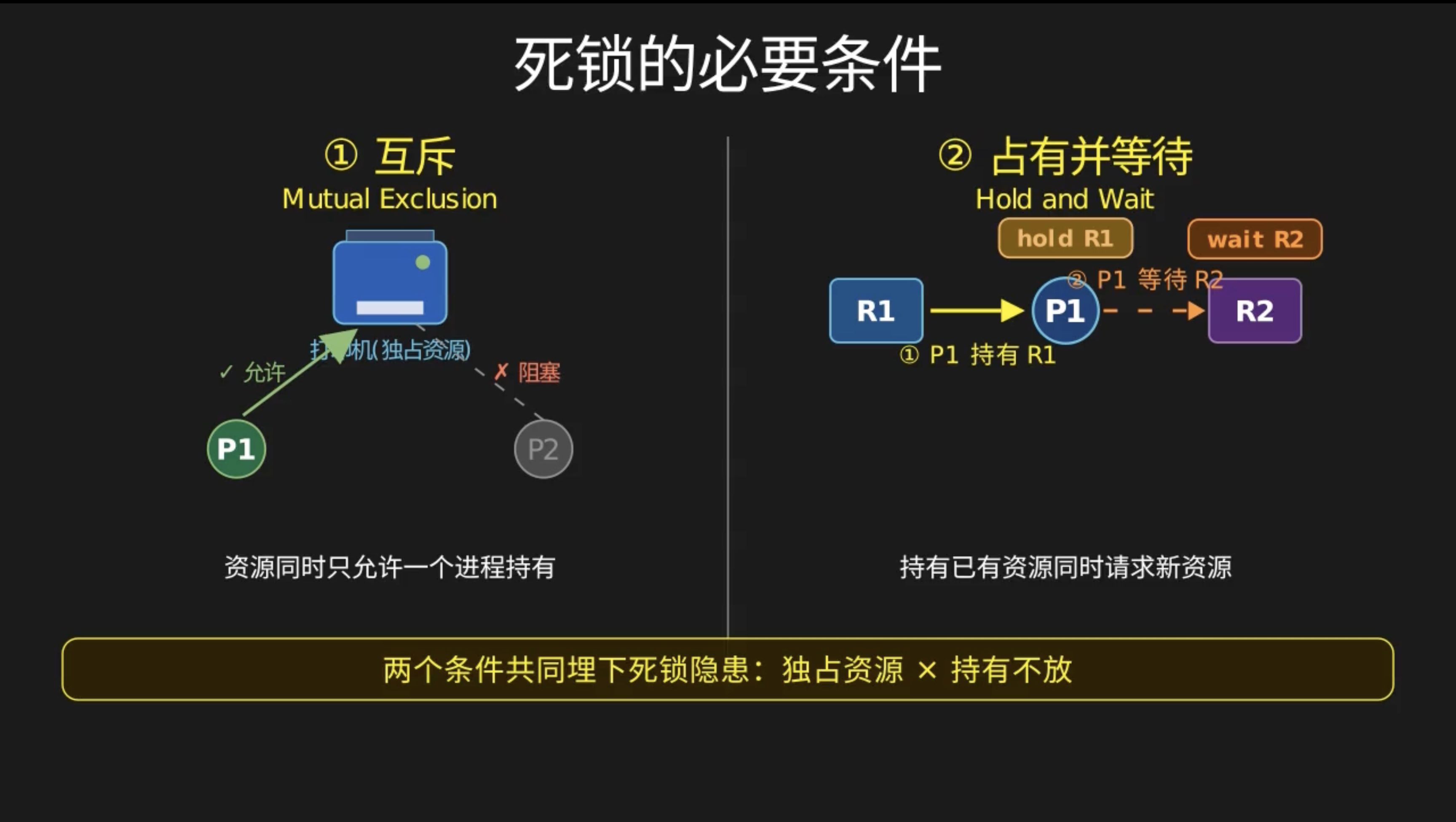}
\end{minipage}

\vspace{0.4em}

\begin{minipage}[t]{0.48\textwidth}
    \centering
    \includegraphics[
        width=\linewidth,
        height=0.18\textheight
    ]{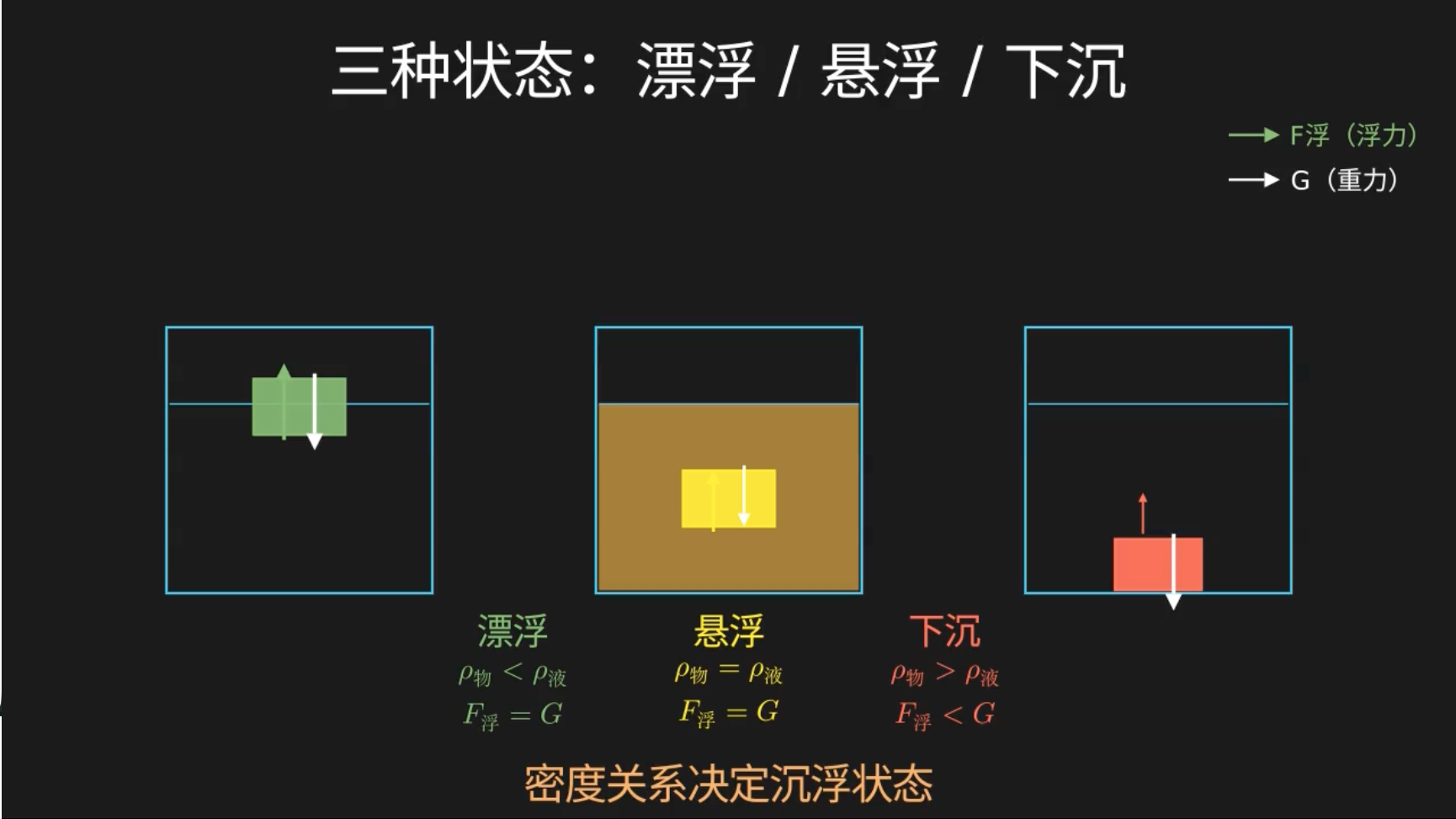}
\end{minipage}
\hfill
\begin{minipage}[t]{0.48\textwidth}
    \centering
    \includegraphics[
        width=\linewidth,
        height=0.18\textheight
    ]{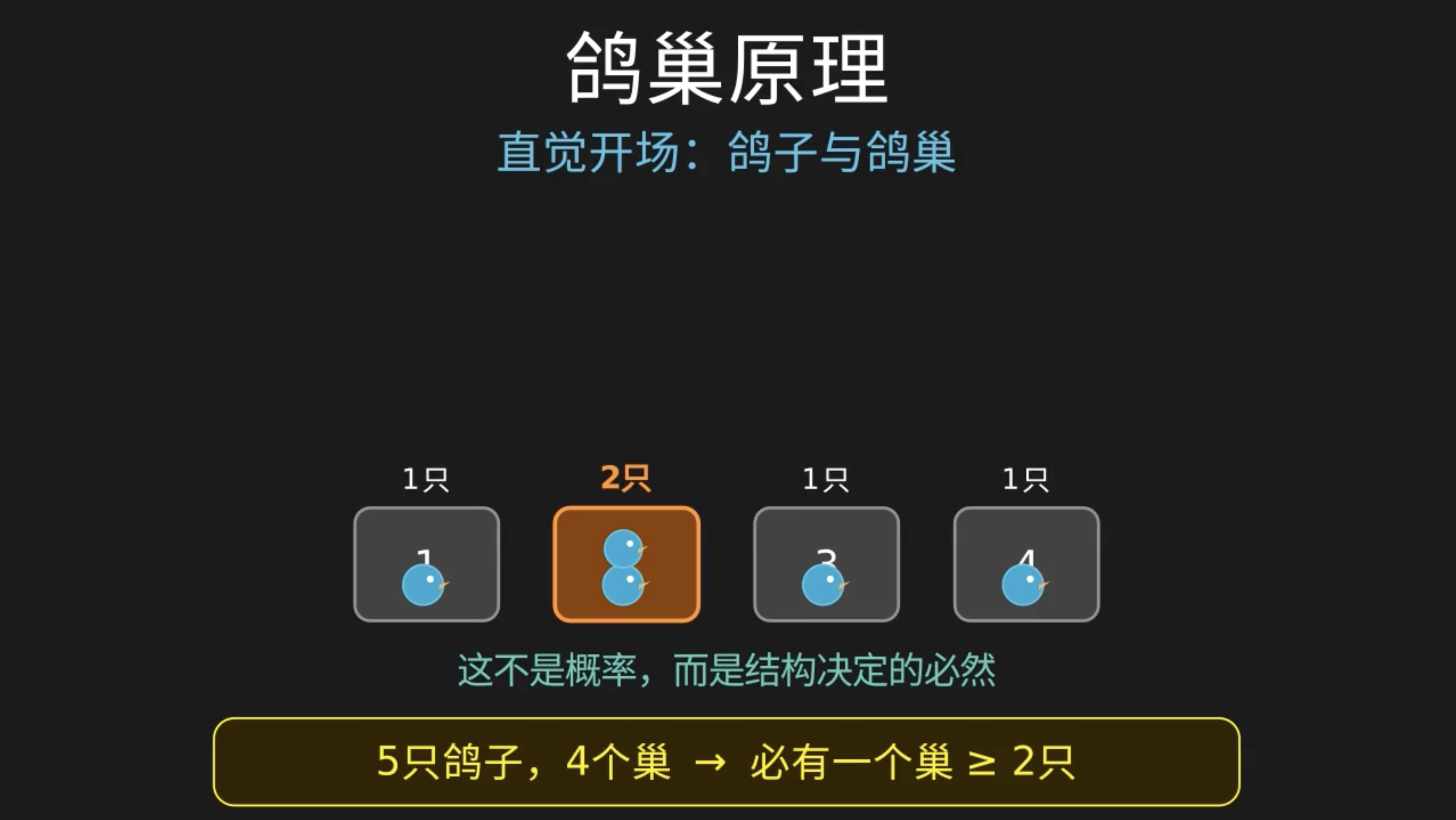}
\end{minipage}

\caption{
Representative Chinese instructional videos generated by PIVOT framework.
}
\label{fig:chinese_case_study}
\end{figure*}

The top row of Figure~\ref{fig:bilingual_case_study} presents a Chinese chemistry example on transition metals, while the bottom row presents an English computer science example on gradient descent optimization. In particular, Figure~\ref{fig:bilingual_case_study}(c) demonstrates that our framework can further integrate externally generated visual assets, where a Nano Banana-generated gradient descent visualization is incorporated into the executable Manim rendering pipeline to improve visual interpretability and instructional expressiveness. Figure~\ref{fig:chinese_case_study} shows additional examples of Chinese instructional videos generated by PIVOT framework.

\end{document}